\documentclass[logo,copyright]{msas}

\usepackage{csquotes}

\usepackage[backend=bibtex,style=alphabetic,minnames=2,maxnames=5,maxcitenames=2,maxbibnames=99,natbib=true]{biblatex}

\usepackage{amsmath}
\usepackage{amsthm}
\usepackage[ruled]{algorithm2e}
\usepackage{tikz}
\usepackage{mathdots}
\usepackage{yhmath}
\usepackage{cancel}
\usepackage{color}
\usepackage{siunitx}
\usepackage{array}
\usepackage{multirow}
\usepackage{amssymb}
\usepackage{gensymb}
\usepackage{tabularx}
\usepackage{nicematrix}
\usepackage{enumitem}
\usepackage{extarrows}
\usepackage{caption}
\usepackage{subcaption}
\usepackage{graphicx}
\usepackage{makecell}
\usepackage{float}
\usepackage{booktabs}
\usepackage{cuted}
\usepackage{capt-of}
\usepackage[capitalise]{cleveref}
\usetikzlibrary{fadings}
\usetikzlibrary{patterns}
\usetikzlibrary{shadows.blur}
\usetikzlibrary{shapes}

\definecolor{high}{HTML}{ef3b2c}  %
\definecolor{low}{HTML}{fff7bc}  %

\crefname{appsec}{Appendix Section}{Appendix Sections}
\crefname{appfig}{Appendix Figure}{Appendix Figures}
\crefname{apptab}{Appendix Table}{Appendix Tables}
\crefname{appeq}{Appendix Equation}{Appendix Equations}

\definecolor{darkpastelgreen}{rgb}{0.01, 0.75, 0.24}

\title{Physics-Assisted and Topology-Informed Deep Learning for Weather Prediction}

\author[1]{Jiaqi Zheng}
\author[1]{Qing Ling}
\author[2]{Yerong Feng}

\affil[1]{Sun Yat-Sen University}
\affil[2]{Shenzhen Institute of Meteorological Innovation}

\newcommand{\name}[1]{PASSAT}

\correspondingauthor{zhengjq23@mail2.sysu.edu.cn, lingqing556@mail.sysu.edu.cn, yerong\_feng@gd121.cn}

\begin{abstract}
Although deep learning models have demonstrated remarkable potential in weather prediction, most of them overlook either the \textbf{physics} of the underlying weather evolution or the \textbf{topology} of the Earth’s surface. In light of these disadvantages, we develop PASSAT, a novel Physics-ASSisted And Topology-informed deep learning model for weather prediction. PASSAT attributes the weather evolution to two key factors: (i) the advection process that can be characterized by the advection equation and the Navier-Stokes equation; (ii) the Earth-atmosphere interaction that is difficult to both model and calculate. PASSAT also takes the topology of the Earth's surface into consideration, other than simply treating it as a plane. With these considerations, PASSAT numerically solves the advection equation and the Navier-Stokes equation on the spherical manifold, utilizes a spherical graph neural network to capture the Earth-atmosphere interaction, and generates the initial velocity fields that are critical to solving the advection equation from the same spherical graph neural network.
In the $5.625^\circ$-resolution ERA5 data set, PASSAT outperforms both the state-of-the-art deep learning-based weather prediction models and the operational numerical weather prediction model IFS T42. 
Code and checkpoint are available at \url{https://github.com/Yumenomae/PASSAT_5p625}.
\end{abstract}

\begin{document}
\maketitle

\section{Introduction}
\label{sec:intro}

\begin{figure}[tb]
    \centering
    \includegraphics[width=30pc]{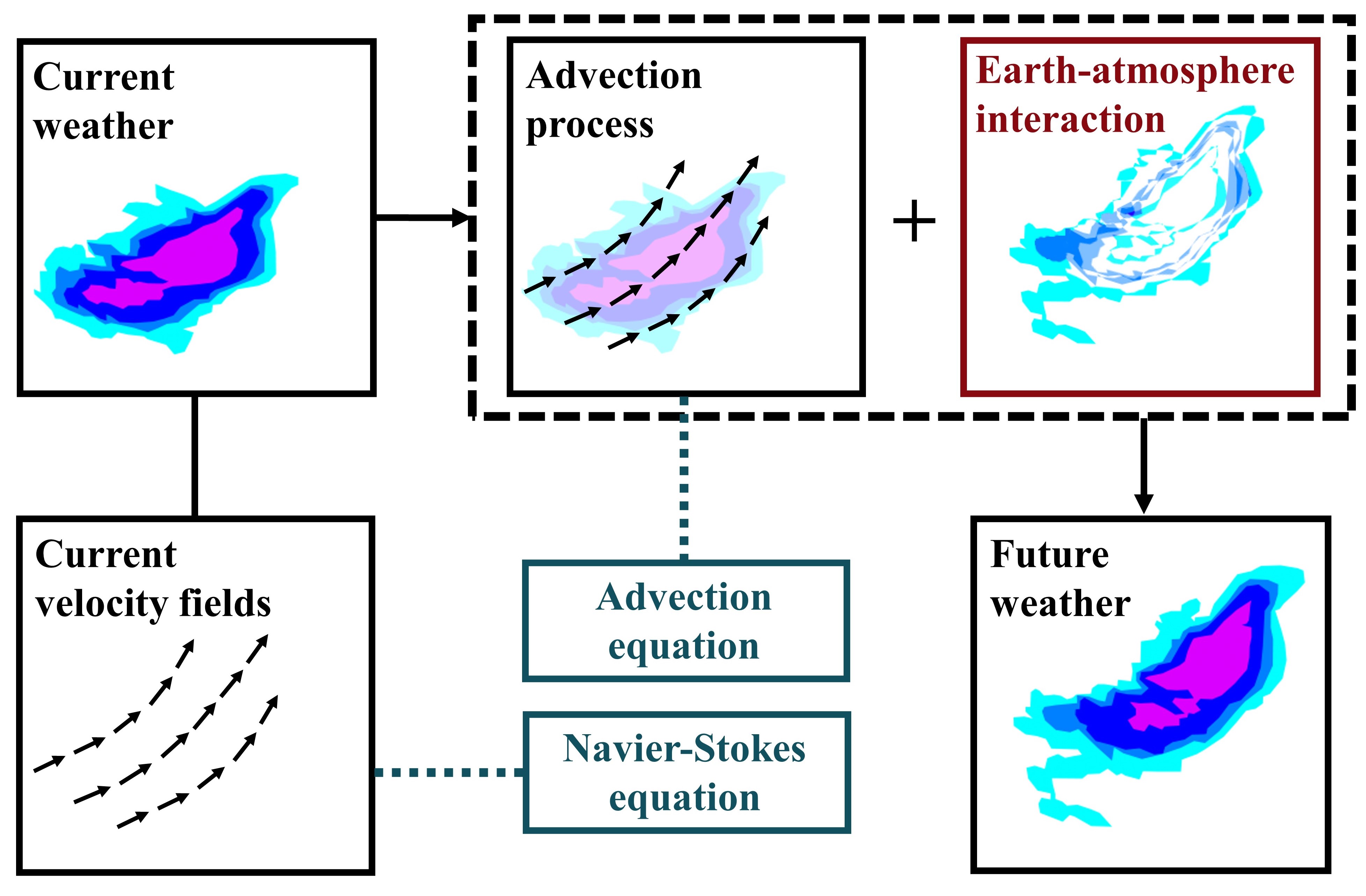}
\caption{\textbf{Attributions of the weather evolution.}}
\label{figure_profor}
\end{figure}

Weather prediction is of paramount importance to social security and economic development, and has attracted extensive research efforts since the ancient time. Among the modern weather prediction methods, numerical weather prediction (NWP) is built upon differential equations that govern the weather evolution \citep{randall2007climate,bauer2015quiet}.
These differential equations attribute the weather evolution to the \textbf{advection process} and the \textbf{Earth-atmosphere interaction} \citep{rood1987numerical,https://doi.org/10.1029/JC095iC06p09461}, as shown in Figure \ref{figure_profor}. The advection process is the evolution of weather variables (described by the advection equation) driven by the evolution of their velocity fields (described by the Navier-Stokes equation) \citep{temam2024navier}. The Earth-atmosphere interaction encompasses other complex physical processes in the atmosphere, such as radiation, clouds, and subgrid turbulent motions. One particular challenge in NWP is that the Earth-atmosphere interaction is difficult to model and calculate, forming a bottleneck of improving the accuracy of NWP \citep{TheArtandScienceofClimateModelTuning,Kochkov2024}. Besides, the accuracy of NWP does not improve with the increasing amount of historical observations.

On the other hand, data-driven methods that predict the weather based on the historical observations, especially deep learning models, have become very popular in recent years \citep{Bouall2024TheRiseofDataDriven}. With the aid of high-quality and ever-accumulating data, state-of-the-art deep learning models have demonstrated great potentials and been integrated into the modern weather prediction systems \citep{10.1145/3592979.3593412,bi2023accurate,doi:10.1126/science.adi2336}. In addition, deep learning-based models are able to remarkably shorten the time consumption in the prediction stage \citep{10.1145/3592979.3593412}. However, these models disregard either the \textbf{physics} of the weather evolution or the \textbf{topology} of the Earth's surface. Thus, their predictions are often unreliable due to the lack of the physical constraints or suffer from the distortions caused by the topological structure \citep{schultz2021can,Xu_2024}.

\subsection{Enhancing Deep Learning with Physics}

Combining with the differential equations that characterize the weather evolution can enhance the precisions, efficiency and robustness of deep learning models, because the differential equations provide valuable prior knowledge \citep{xiang2022self}. Some works incorporate differential equations into losses during training deep learning models \citep{daw2021physicsguidedneuralnetworkspgnn}. Nevertheless, tuning weights for the differential equations and computing stochastic gradients of the losses bring new challenges. Some other works use deep learning models to correct NWP models \citep{https://doi.org/10.1029/2022MS003400,xu2024generalizingweatherforecastfinegrained,Kochkov2024}. Though having high accuracies, these approaches are computationally demanding since they need to both solve a large system of differential equations and train end-to-end neural networks. The closest to ours are \citep{zhang2023skilful,verma2024climode}, in which neural networks are trained with the aid of differential equations. However, they both overlook the Navier-Stokes equation that drives the evolution of the velocity fields.

Despite that these physics-assisted deep learning models are harder to train and slower in inference compared to the end-to-end deep learning methods, they significantly enhance the robustness of predictions and demonstrate remarkable potentials \citep{chen2018neural}.

\subsection{Taking Topology of Earth into Consideration}

\begin{figure}[tb]
    \centering
    \includegraphics[width=30pc]{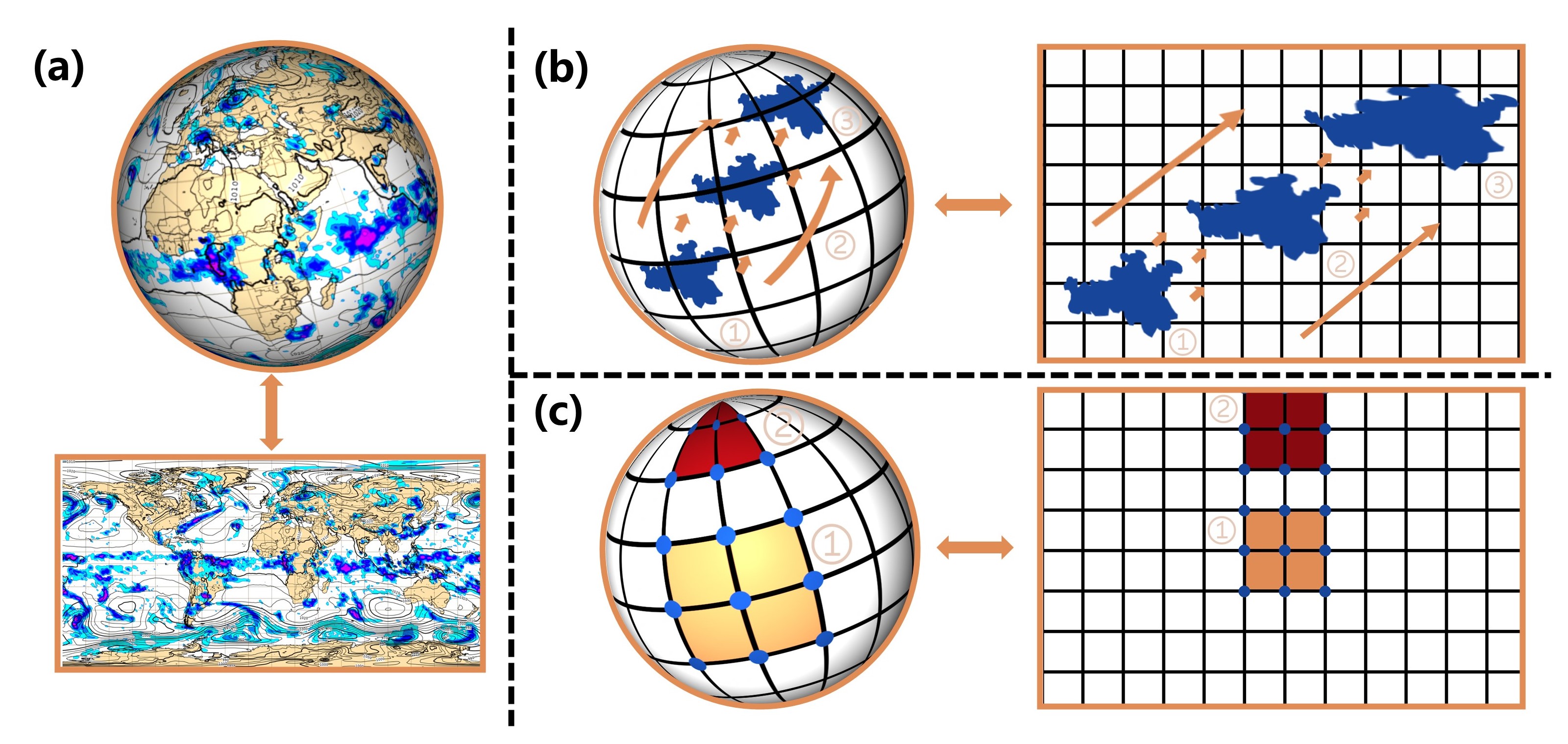}
\caption{\textbf{Distortions due to planar projection.}
(a) The spherical and planner representations of the global weather.
(b) The same weather patterns on the sphere are distorted on the plane.
(c) The convolutions on the sphere are distorted on the plane.}
\label{figure_topology}
\end{figure}

The historical observations used during training most deep learning-based weather prediction models are often on planar latitude-longitude grids, other than on the spherical surface of the Earth. However, neglecting the topology of the Earth's surface introduces remarkable distortions, as shown in Figure \ref{figure_topology} \citep{s.2018spherical,mai2023sphere2vec}. For example, the points that are close to the poles turn to be denser on the spherical manifold than on the planar latitude-longitude grid. A notable consequence is that one weather pattern appears differently on the sphere and the plane, such that capturing the weather pattern on the plan suffers from distortions. These distortions also affect the patches and convolution kernels, negatively impacting the deep learning models based on convolutional neural networks or transformers \citep{coors2018spherenet}. In addition, the velocity fields defined on the planar are significantly distorted when increasing the latitude towards the poles, which will bring biases to the deep learning models that learn the velocity fields \citep{zhang2023skilful,verma2024climode}.

\subsection{Contributions}

In this paper, we propose PASSAT, a novel \textbf{P}hysics-\textbf{ASS}isted \textbf{A}nd \textbf{T}opology-informed deep learning model for weather prediction. PASSAT attributes the weather evolution to the analytical advection process and the complex Earth-atmo- sphere interaction. In the advection process, the evolution of weather variables is driven by the evolution of their velocity fields, and the two are respectively described by the advection equation and the Navier-Stokes equation. PASSAT also takes the topology of the Earth's surface into consideration. Therefore, PASSAT: (i) trains a spherical graph neural network to estimate the Earth-atmosphere interaction; (ii) generates the initial velocity fields with the same spherical graph neural network; (iii) numerically solves the advection equation on the spherical manifold; (iv) updates the velocity fields through numerically solving the Navier-Stokes equation on the spherical manifold. Our contributions are as follows.

\begin{itemize}
\item PASSAT seamlessly integrates the historical observations, the physics of the weather evolution and the topology of the Earth's surface, yielding a novel physics-assisted and topology-informed deep learning model for weather prediction.
\item Compared to the black-box deep learning models, PASSAT takes advantages of the physical constraints, characterized by the advection equation and the Navier-Stokes equation, and thus remarkably improves the quality of medium-term prediction.
\item Compared to the NWP models, PASSAT avoids modeling and calculating the complex Earth-atmosphere interaction. PASSAT is also able to utilize the historical observations to improve the prediction accuracy.
\item PASSAT solves the differential equations and trains the graph neural network on the spherical manifold other than on the planar latitude-longitude grid, and thus effectively avoids the distortions brought by the latter.
\item We conduct experiments on the 5.625$^\circ$ ERA5 data set, demonstrating the competitive performance of PASSAT compared to the state-of-the-art deep learning models and the NWP model IFS T42.
\end{itemize}

\section{Related Works}

\noindent \textbf{Numerical weather prediction (NWP).} NWP is a fundamental physics-based method for weather prediction \citep{https://doi.org/10.1029/2018GL080704}, utilizing the underlying differential equations to predict how the weather should evolve over the time. For example, the operational Integrated Forecast System (IFS) consists of several NWP models with different spatial resolutions \citep{Bouall2024TheRiseofDataDriven}. Despite of its widespread applications, modeling and calculating the complex Earth-atmosphere interaction are challenging. In addition, solving the differential equations is sensitive to the initial conditions, and also computationally demanding \citep{Kochkov2024}.


\noindent \textbf{Deep learning-based weather prediction.} Different from NWP, deep learning models learn from the historical observations to predict the weather. Though time-consuming during training, deep learning models are rapid during prediction as they do not involve solving the differential equations. State-of-the-art deep learning-based weather prediction models include FourCastNet \citep{10.1145/3592979.3593412}, SFNO \citep{bonev2023sphericalfourierneuraloperators}, GraphCast \citep{doi:10.1126/science.adi2336}, Graph-EFM \citep{keisler2022forecastingglobalweathergraph}, AIFS \citep{lang2024aifsecmwfsdatadriven}, Pangu \citep{bi2023accurate}, Fengwu \citep{chen2023fengwu}, Fuxi \citep{chen2023fuxi}, Stormer \citep{nguyen2024scalingtransformerneuralnetworks}, etc. FourCastNet and SFNO are based on the Fourier neural operator \citep{li2021fourierneuraloperatorparametric}, GraphCast, Graph-EFM and AIFS are based on the graph neural network \citep{9046288}, whereas Pangu, Fengwu, Fuxi, and Stormer utilize the vision \citep{dosovitskiy2021} and swin \citep{liu2021swintransformerhierarchicalvision} transformers. Among these models, SFNO, GraphCast, Graph-EFM, and AIFS take the Earth's topology into consideration. Nevertheless, all of them disregard the underlying physics information.

\noindent \textbf{Deep learning-based, physics-assisted weather prediction.} Integrating the differential equations with deep learning models significantly improves the precisions, efficiency and robustness of the latter. Notable recent works along this line include ClimODE \citep{verma2024climode} and NowcastNet \citep{zhang2023skilful}. Different to PASSAT, ClimODE characterizes the evolution of the weather variables with the continuity equation, other than the advection equation. On the other hand, ClimODE updates the velocity fields with a neural network, other than the Navier-Stokes equation. NowcastNet focuses on regional precipitation nowcasting, while PASSAT focuses on global, multi-variable and medium-term weather prediction. Besides, PASSAT solves the differential equations and trains its graph neural network on a spherical manifold, other than on the planar latitude-longitude grid used by ClimODE and NowcastNet, effectively avoiding the distortions.
\section{Methods}

\begin{figure}[tb]
    \centering
    \includegraphics[width=30pc]{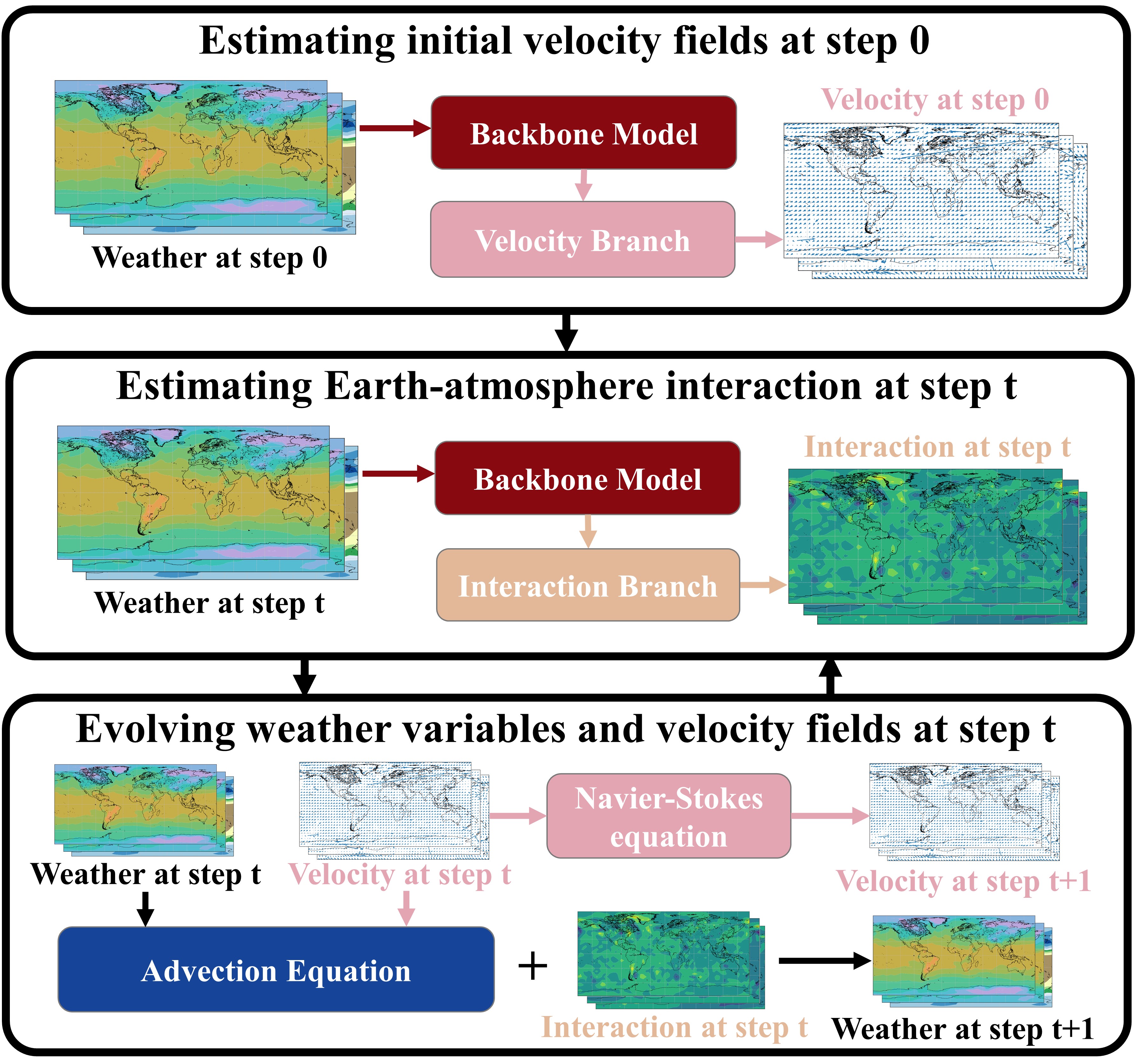}
	\caption{\textbf{Overview of PASSAT.}}
\label{figure_ov}
\end{figure}

Considering the attributions of the weather evolution demonstrated in Figure \ref{figure_profor}, we accordingly build a physics-assisted and topology-informed deep learning model for weather prediction, abbreviated as PASSAT. Given any initial time, PASSAT: (i) generates the initial velocity fields of the weather variables with the velocity branch of a spherical graph neural network; and then autoregressively (ii) predicts the effects of the Earth-atmosphere interaction with the interaction branch of the spherical graph neural network; (iii) numerically solves the advection equation on the spherical manifold; (iv) numerically updates the velocity fields through solving the Navier-Stokes equation on the spherical manifold, aided by the initial velocity fields provided by (i). In the following, we discuss how PASSAT captures the evolution of the weather variables and their velocity fields, via integrating the two differential equations and the spherical graph neural network (see also Figure \ref{figure_ov}).

We disregard the impact of vertical actions and focus on analyzing the advection and Navier-Stokes equations on the spherical manifold. All the analyses and approaches presented below can be readily extended to scenarios where the vertical actions are taken into account.

We begin by introducing the spherical manifold in Section \ref{snn} and describing the evolution of the weather variables in Section \ref{obj}. Then, we present the advection equation on the spherical manifold, the spherical graph neural network and the Navier-Stokes equation on the spherical manifold in Section  \ref{adv}, Section \ref{gnn} and Section \ref{NS}, respectively. We also introduce the time integration scheme in Section \ref{tis}. Finally, we summarize in Section \ref{sum}.

\subsection{Spherical Manifold} \label{snn}

The historical observations used during training most deep learning-based weather prediction models are often on planar latitude-longitude grids, other than on the spherical surface of the Earth. Ignoring this topology information leads to remarkable distortions in both the neural networks and the differential equations \citep{s.2018spherical,mai2023sphere2vec}. In order to avoid such distortions, we project the weather variables from a planar latitude-longitude grid onto the Earth's surface. We assume the Earth's surface to be an ideal unit sphere, with the radius of 1 unit length (6371km).

We denote the unit sphere $\mathbf{S} = \{\mathbf{s} \in \mathbb{R}^{3} | \ ||\mathbf{s}||_{2}=1\}$ as the Earth's surface. Any spatial coordinate $\mathbf{s}$ on the unit sphere corresponds to a point $(\phi,\theta)$ within the planar latitude-longi- tude grid, where $\theta$ is the latitude and $\phi$ is the longitude. Thus, we use $\mathbf{s}$ and $\mathbf{s}(\phi,\theta)$ interchangeably. Given any spatial coordinate $\mathbf{s}$, $\mathbf{e}_{\phi}(\mathbf{s}) \in \mathbb{R}^{3}$ and $\mathbf{e}_{\theta}(\mathbf{s}) \in \mathbb{R}^{3}$ are two orthogonal unit vectors originated from $\mathbf{s}$ and along the parallel and meridian directions, respectively. We denote $\nabla_{\mathbf{s}}$ as the spatial gradient on the unit sphere and $\cdot$ as the inner product.

\subsection{Evolution of Weather Variables} \label{obj}

Weather prediction depends on understanding the evolution of weather variables that we are interested in. Given any wea- ther variable $u$, its evolution is characterized as follows.

The weather variable $u$ is viewed as a differentiable, real-valued function  $u: T \times \mathbf{S} \rightarrow \mathbb{R}$, within which $T$ is the time set and $\mathbf{S}$ is the Earth's surface. According to Figure \ref{figure_profor}, the evolution of weather variable $u$ is attributed to the advection process and the Earth-atmosphere interaction. To be specific, for any $(t, \mathbf{s})$ $\in T \times \mathbf{S}$, we have:
\begin{align} \label{att}
	u(t+\delta t, \mathbf{s} + \delta \mathbf{s}) \approx  u(t, \mathbf{s}) + \int_{t}^{t+\delta t}  h(\tau, \mathbf{s} + \delta \mathbf{s})d\tau,
\end{align}
where $\delta t > 0$ is a small lead time, $\delta \mathbf{s} = \delta t \times \mathbf{v}(t, \mathbf{s})$, $\mathbf{v}$ is the velocity fields of $u$, and $h$ is the tendency of $u$ due to the Earth-atmosphere interaction. Taking the first-order Taylor's approximation of (\ref{att}) yields:
\begin{align} \label{att2}
	\frac{\partial u}{\partial t}(t, \mathbf{s}) \approx  -  \mathbf{v}(t, \mathbf{s})  \cdot \nabla_{\mathbf{s}}u(t, \mathbf{s}) + h(t, \mathbf{s})  = (\frac{\partial u}{\partial t}(t, \mathbf{s}))_{\textbf{advection}} + (\frac{\partial u}{\partial t}(t, \mathbf{s}))_{\textbf{interaction}}.
\end{align}

According to \eqref{att2}, the \textbf{total tendency} of weather variable $u$ can be decomposed into two part: (i) $ - \mathbf{v} \cdot \nabla_{\mathbf{s}}u$, the tendency due to the \textbf{advection process} and (ii) $h$, the tendency due to the \textbf{Earth-atmosphere interaction}. Once the total tendency $\frac{\partial u}{\partial t}(t, \mathbf{s})$ is known, we can predict the value of $u$ at any future time $t+\Delta t$ via using proper numerical methods to solve:
\begin{align} \label{update}
	u(t + \Delta t, \mathbf{s}) &= u(t, \mathbf{s}) + \int_{t}^{t+\Delta t} \frac{\partial u}{\partial t}(\tau, \mathbf{s}) d\tau.
\end{align}
In PASSAT, we use Euler's method for this time integration.

Therefore, the key of weather prediction is to compute the tendencies of the advection process and the Earth-atmosphere interaction. Though the tendency of the advection process can be numerically estimated by solving the advection equation on the spherical manifold, the tendency of the Earth-atmosphere interaction is difficult to model and calculate so that we resort to a spherical graph neural network. We introduces them one by one in the following.

\subsection{Advection Equation on Spherical Manifold} \label{adv}

The advection process is the evolution of the weather variables driven by their velocity fields. Given any weather variable $u$, its velocity fields $\mathbf{v}: T \times \mathbf{S} \rightarrow \mathbb{R}^{3}$ are differentiable functions of time and spatial coordinate. As we disregard vertical actions, the velocity fields can be express by $\mathbf{v}(t,\mathbf{s}) = $ $v_{\theta}(t,\mathbf{s})\mathbf{e}_{\theta}(\mathbf{s}) + v_{\phi}(t,\mathbf{s})\mathbf{e}_{\phi}(\mathbf{s})$, where $v_{\theta}$ and $v_{\phi}$ are the velocities of $u$ along the meridian and parallel directions, respectively. With particular note, at any initial time $t$ and spatial coordinate $\mathbf{s}$, $u(t,\mathbf{s})$ is known but $\mathbf{v}(t,\mathbf{s})$ is to be calculated.

As discussed in Section \ref{obj}, the tendency of $u$ due to the advection process is given by solving the advection equation \citep{Chandrasekar_2022}, as:
\begin{align} \label{advectionequation}
	(\frac{\partial u}{\partial t}(t, \mathbf{s}))_{\textbf{advection}} + & \underbrace{\mathbf{v}(t,\mathbf{s}) \cdot \nabla_{\mathbf{s}} u(t, \mathbf{s})}_{\text{advective derivative}} = 0.
\end{align}
Once the advective derivative is known, the tendency of $u$ due to the advection process is known too. On the spherical manifold and the planar latitude-longitude grid, the advective derivative has different forms, and the latter brings distortions in weather prediction, as discussed in the following.

Given a spatial coordinate $\mathbf{s}=\mathbf{s}(\phi,\theta) \in \mathbf{S}$, on the spherical manifold, the advective derivative is in the form of \citep{lions1992new}:
\begin{equation} \label{advective}
\hspace{-0.2em} \mathbf{v}(t,\mathbf{s}) \cdot \nabla_{\mathbf{s}} u(t, \mathbf{s}) =  v_{\theta}(t, \mathbf{s})\frac{\partial u}{\partial \theta}(t, \mathbf{s}) + \frac{v_{\phi}(t, \mathbf{s})}{\cos\theta}\frac{\partial u}{\partial \phi}(t, \mathbf{s}).
\end{equation}
For $v_{\theta}(t, \mathbf{s})$ and $v_{\phi}(t, \mathbf{s})$, PASSAT will estimate their initial values utilizing the velocity branch of a spherical graph neural network, and calculate their future values through solving the Navier-Stokes equation. The differentials $\frac{\partial u}{\partial \theta}(t, \mathbf{s})$ and $\frac{\partial u}{\partial \phi}(t, \mathbf{s})$ can be estimated using the difference quotients of $u$ on the planar latitude-longitude grid.

In contrast, on the planar latitude-longitude grid, the advective derivative is in the form of:
\begin{equation}\label{eq5}
\mathbf{v}(t,\mathbf{s}) \cdot \nabla_{\mathbf{s}} u(t, \mathbf{s}) =  v'_{\theta}(t,\mathbf{s})\frac{\partial u}{\partial \theta}(t,\mathbf{s}) + v'_{\phi}(t,\mathbf{s}) \frac{\partial u}{\partial \phi}(t,\mathbf{s}),
\end{equation}
where $v'_{\theta}(t, \mathbf{s})$ and $v'_{\phi}(t, \mathbf{s})$ are respectively the velocities along the meridian and parallel directions, but on the latitude-longitude planar grid, not on the spherical manifold. We have $v'_{\theta}(t,\mathbf{s}) = v_{\theta}(t,\mathbf{s})$ and $v'_{\phi}(t,\mathbf{s}) = \frac{v_{\phi}(t, \mathbf{s})}{\cos\theta}$.

ClimODE and NowcastNet both calculate the the advective derivative according to (\ref{eq5}), through estimating $v'_{\theta}(t, \mathbf{s})$ and $v'_{\phi}(t, \mathbf{s})$ with neural networks. However, we can observe that fixing the value of $v_{\phi}(t, \mathbf{s})$, $v'_{\phi}(t, \mathbf{s})$ is not spatial-invariant -- it is large when $\mathbf{s}$ is close to the poles and small when $\mathbf{s}$ is close to the equator. Such distortions will affect the pattern recognition of the neural networks. In contrast, PASSAT takes advantages of the spherical manifold, and thus avoids the distortions.

\subsection{Spherical Graph Neural Network}\label{gnn}

\begin{figure}[tb]
    \centering
	\hspace{-7.5mm}
    \includegraphics[width=30pc]{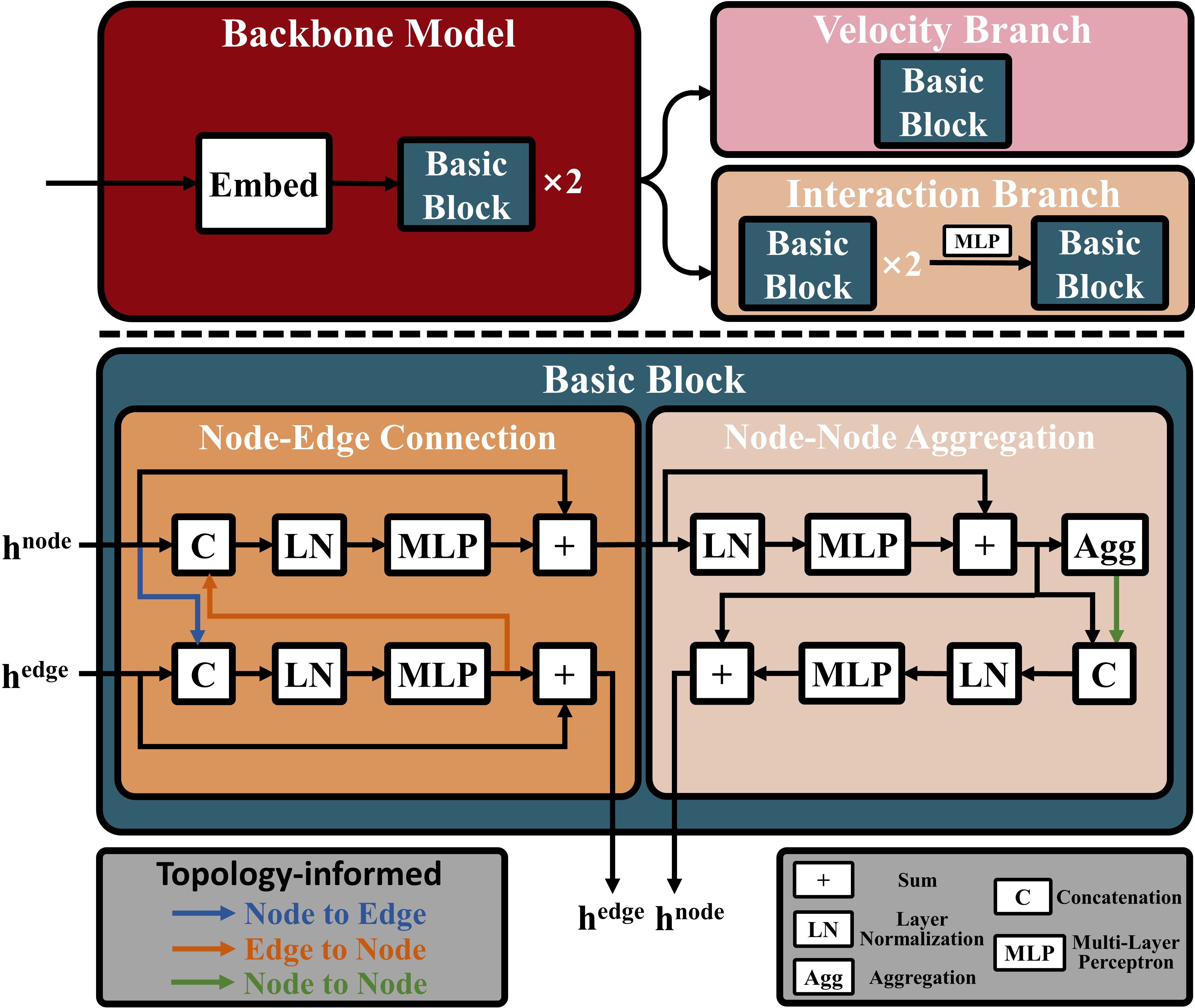}
	\caption{\textbf{Overview of PASSAT's graph neural network.} TOP: The backbone model and two branches. BOTTOM: The basic block.}
\label{figure_gnn}
\end{figure}

As discussed above, to calculate $(\frac{\partial u}{\partial t}(t, \mathbf{s}))_{\textbf{advection}}$, the tendency of $u$ due to the advection process, we need to estimate the initial velocity fields of $\mathbf{v}(t,\mathbf{s})$. On the other hand, we also need to estimate $(\frac{\partial u}{\partial t}(t, \mathbf{s}))_{\textbf{interaction}}$, the tendency of $u$ due to the Earth-atmosphere interaction. We train a spherical graph neural network to estimate these values.

The spherical graph neural network consists of a backbone model along with two branches: the interaction branch that estimates $(\frac{\partial u}{\partial t}(t, \mathbf{s}))_{\textbf{interaction}}$ and the velocity branch that estimates the initial velocity fields of $\mathbf{v}(t,\mathbf{s})$; see Figure \ref{figure_gnn}. Their basic block has two components: \textbf{node-edge connection} and \textbf{node-node aggregation}. The node-edge connection component enables efficient message passing between nodes and edges \citep{ZHOU202057}, while the node-node aggregation component performs graph convolution among nodes \citep{kipf2016semi}.

The spherical graph neural network incorporates the topology information from the spherical manifold, and thus avoids the distortions caused by the planar latitude-longitude grid. For more details, please refer to the supplementary material.

\subsection{Navier-Stokes Equation on Spherical Manifold} \label{NS}
To calculate the future $(\frac{\partial u}{\partial t}(t, \mathbf{s}))_{\textbf{advection}}$, we still need to estimate the future velocity fields of $\mathbf{v}(t,\mathbf{s})$, for which we resort to the Navier-Stokes equation \citep{lions1992new}.
On the spherical manifold, the velocity fields of each weather variable $\mathbf{v}(t,\mathbf{s}) = v_{\theta}(t,\mathbf{s})\mathbf{e}_{\theta}(\mathbf{s}) + v_{\phi}(t,\mathbf{s})\mathbf{e}_{\phi}(\mathbf{s})$ satisfy:
\begin{align}\label{update_lat_vel}
	\frac{\partial v_{\theta}}{\partial t} + \underbrace{(v_{\theta}\frac{\partial v_{\theta}}{\partial \theta}+\frac{v_{\phi}}{\cos\theta}\frac{\partial v_{\theta}}{\partial \phi})}_{\text{advection}}  + \underbrace{v_{\phi}^{2}\tan\theta}_{\text{curvature}} +  \underbrace{ \frac{1}{\rho }\frac{\partial p}{\partial \theta}}_{\text{pressure gradient force}} +  \underbrace{2\omega v_{\phi}\sin\theta}_{\text{Coriolis force}} + \underbrace{\frac{\mu}{\cos^{2}\theta}v_{\theta}}_{\text{viscous friction}} = 0,
\end{align}

\begin{align}\label{update_lon_vel}
\frac{\partial v_{\phi}}{\partial t} + \underbrace{(v_{\theta}\frac{\partial v_{\phi}}{\partial \theta} + \frac{v_{\phi}}{\cos\theta}\frac{\partial v_{\phi}}{\partial \phi})}_{\text{advection}}  - \underbrace{ v_{\phi}v_{\theta}\tan\theta}_{\text{curvature}} + \underbrace{ \frac{1}{\rho \cos\theta}\frac{\partial p}{\partial \phi}}_{\text{pressure gradient force}} - \underbrace{ 2\omega v_{\theta}\sin\theta}_{\text{Coriolis force}} + \underbrace{\frac{\mu}{\cos^{2}\theta}v_{\phi}}_{\text{viscous friction}} = 0,
\end{align}
We omit the pair $(t,\mathbf{s})$ for notational simplicity. In the Navier-Stokes equation, $\rho(t, \mathbf{s})$ is the atmospheric density, $\omega = 0.2618 \ ( \text{radian/hour})$ is the Earth's rotation speed, $p(t,\mathbf{s})$ is the atmospheric pressure, and $\mu$ is a constant related to the Reynolds constant. For computational efficiency, we simplify the Navier-Stokes equation by retaining only the viscous friction in the Laplacian.

The Navier-Stokes equation governs the evolution of both $v_{\theta}(t,\mathbf{s})$ and $v_{\phi}(t,\mathbf{s})$. After calculating $\frac{\partial v_{\theta}(t,\mathbf{s})}{\partial t}$ and $\frac{\partial v_{\phi}(t,\mathbf{s})}{\partial t}$ from the Navier-Stokes equation, we apply numerical methods to predict $v_{\theta}(t+\Delta t,\mathbf{s})$ and $v_{\phi}(t+\Delta t,\mathbf{s})$ as in \eqref{update}.

\subsection{Time Integration Scheme}\label{tis}

Up to now, at time $t$, we have known $(\frac{\partial u}{\partial t}(t, \mathbf{s}))_{\textbf{interaction}}$ (from the spherical graph neural network) and $(\frac{\partial u}{\partial t}(t, \mathbf{s}))_{\textbf{advection}}$ (from the advection equation) with the aid of $\mathbf{v}(t,\mathbf{s})$ (from both the spherical graph neural network and the Navier-Stokes equation). Therefore, we can the predict the value of $u$ at a future time $t+\Delta t$ according to (\ref{update}). However, the numerical methods to solve (\ref{update}) are sensitive to the integration step size. As we will see, in our weather prediction, $\Delta t$ ranges from 6 to 144 hours, at the temporal resolution of 6 hours. Hence, choosing a proper integration step size is critical to medium-term or long-term prediction.

With the above consideration, in PASSAT, we set the integration step size as $0.2$ hours. However, such a small integration step size requires the interaction branch of PASSAT to frequently estimate $(\frac{\partial u}{\partial t}(t, \mathbf{s}))_{\textbf{interaction}}$ -- see \eqref{att2} and \eqref{update} -- resulting in excessive back propagation during the training. To address the issue of memory-intensive training, we estimate $(\frac{\partial u}{\partial t}(t, \mathbf{s}))_{\textbf{interaction}}$ only once every hour and keep it unchanged within the next hour. 

\subsection{Summary of PASSAT}\label{sum}

We summarize PASSAT in Algorithm \ref{alg1}. For simplicity, we omit the spatial coordinate $\mathbf{s}$, using $u^t$ to denote any weather variable at time $t$ and $\mathbf{v}^t$ to denote its velocity fields. We also use $\mathbf{u}^{t}$ to denote all weather variables at time $t$. We use $f_{\mathbf{vel}}$ and $f_{\mathbf{int}}$ to denote the velocity and interaction branches of the spherical graph neural network, respectively. We use $\mathbf{i}^t$ to denote the estimated effect of the Earth-atmosphere interaction for time $t$ and spatial coordinate $\mathbf{s}$. Within the Navier-Stokes equation, $\frac{1}{\rho} \frac{\partial p^t}{\partial \phi}$ and $\frac{1}{\rho} \frac{\partial p^t}{\partial \theta}$ are unknown. We replace them with the gradients of geopotential $z^t$ at the 500hPa pressure level, converting the units from $\text{m}^{2}\text{second}^{-2}$ to $(6731 \text{km})^{2}\text{hour}^{-2}$. To ensure stability, the initial velocity fields estimated from the velocity branch, with the unit of $(6731\text{km})/\text{hour}$, are projected onto $[-0.005, 0.005]$.

\begin{algorithm}[tb]
        \caption{PASSAT: Predicting any weather variable $u$ for $\tau = t+0.2, t+0.4, \cdots, t+\Delta t$ at time $t$}
        \KwIn{$u^t$}
        \KwOut{$u^{\tau}$ and $\mathbf{v}^{\tau} = v_{\theta}^{\tau}\mathbf{e}_{\theta} + v_{\phi}^{\tau}\mathbf{e}_{\phi}$}
        \For{$z=t,t+1,\cdots,t+\Delta t-1$}
            {
            \If{$z=t$}
            {
               Initial velocity fields: $ \mathbf{v}^t  = f_{\mathbf{vel}}(\mathbf{u}^t )$ \\
            }
		Earth-atmosphere interaction: $\mathbf{i}^z = f_{\mathbf{int}}(\mathbf{u}^{z})$ \\	
		\For{$\tau=z, z+0.2, z+0.4, z+0.6, z+0.8$}{
		---Compute tendencies of $u, v_{\theta}, v_{\phi}$--- \\
		$\left\{\begin{array}{l}
		\frac{\partial u^\tau}{\partial \tau} = - v_{\theta}^{\tau}\frac{\partial u^\tau}{\partial \theta} - \frac{v_{\phi}^{\tau}}{\cos\theta} \frac{\partial u^\tau}{\partial \phi} + \mathbf{i}^z \\

		\frac{\partial v_{\theta}^{\tau}}{\partial \tau} =  - v_{\theta}^{\tau}\frac{\partial v_{\theta}^{\tau}}{\partial \theta} - \frac{v_{\phi}^{\tau}}{\cos\theta} \frac{\partial v_{\theta}^{\tau}}{\partial \phi} -  (v_{\phi}^{\tau})^{2}\tan \theta- \frac{\partial z^\tau}{\partial \theta}-2\omega v_{\phi}^{\tau} \sin\theta-\mu \frac{v_{\theta}^{\tau}}{\cos^{2}\theta} \\

		\frac{\partial v_{\phi}^{\tau}}{\partial \tau} = - v_{\theta}^{\tau}\frac{\partial v_{\phi}^{\tau}}{\partial \theta} - \frac{v_{\phi}^{\tau}}{\cos\theta} \frac{\partial v_{\phi}^{\tau}}{\partial \phi} + v_{\phi}^{\tau} v_{\theta}^{\tau} \tan \theta -  \frac{1}{\cos \theta} \frac{\partial z^\tau}{\partial \phi} + 2 \omega v_{\theta}^{\tau} \sin \theta - \mu \frac{v_{\phi}^{\tau}}{\cos^{2}\theta} \\
		\end{array}\right.$ \\
	\ \\
		---------Update $u^\tau, v_{\theta}^{\tau}, v_{\phi}^{\tau}$--------- \\
$\left\{\begin{array}{l}
u^{\tau+0.2} = u^\tau + 0.2 \times \frac{\partial u^\tau}{\partial \tau} \\
v_{\theta}^{\tau+0.2} = v_{\theta}^{\tau} + 0.2 \times \frac{\partial v_{\theta}^{\tau}}{\partial \tau} \\
v_{\phi}^{\tau+0.2} = v_{\phi}^{\tau} + 0.2 \times \frac{\partial v_{\phi}^{\tau}}{\partial \tau}
\end{array}\right.$
    }}
	\label{alg1}
\end{algorithm}
\section{Experiments}

\begin{figure}[tb]
    \centering
    \includegraphics[width=40pc]{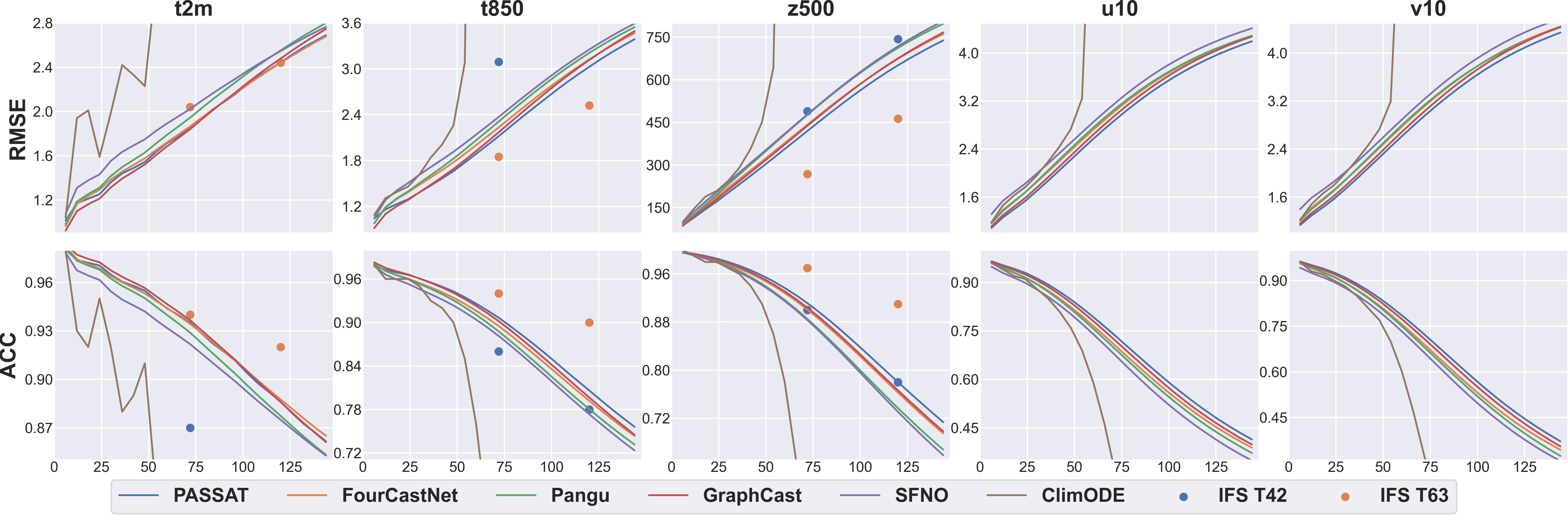}
	\caption{\textbf{Comparison between PASSAT and other models.} The x-axis represents the lead time in hours. Smaller RMSE and larger ACC values indicate better performance. Note that some results of IFS T42 exceed the bounds.}
\label{figure_rmse}
\end{figure}

\noindent \textbf{Data \& Tasks.} We conduct the experiments on the European Centre for Medium-Range Weather Forecasts Reanalysis V5 (ERA5) $5.625^\circ$-resolution data set, spanning from 1979 to 2018 and provided by WeatherBench \cite{hersbach2020era5,rasp2020weatherbench}. The data samples from 1979 to 2015 are used in the training set, 2016 in the validation set, as well as 2017 and 2018 in the test set. The interested weather variables are temperature at 2m height (t2m), temperature at 850hPa pressure level (t850), geopotential at 500hPa pressure level (z500), u component of wind at 10m height (u10), and v com- ponent of wind at 10m height (v10).

We use PASSAT and the baseline models to predict these weather variables, at a temporal resolution of 6 hours (6am, 12am, 6pm, and 12pm of each day) and lasting for 24 steps (144 hours). Performance metrics include root mean square error (RMSE) and anomaly correlation coefficient (ACC).

\noindent \textbf{Baseline deep learning models.} We compare PASSAT with the following baseline deep learning models: (i) ClimODE \cite{verma2024climode};
(ii) FourCastNet \cite{10.1145/3592979.3593412};
(iii) Pangu \cite{bi2023accurate};
(iv) GraphCast \cite{doi:10.1126/science.adi2336};
(v) SFNO \cite{bonev2023sphericalfourierneuraloperators}.
For fair comparisons, we unify the number of parameters of all models to the same magnitude (around 1.15 million) and train these baseline deep learning models from scratch according to their open-source codes and NVIDIA's Modulus\footnote{https://github.com/NVIDIA/modulus/tree/main/modulus/models}.

We do not compare with NowcastNet, Graph-EFM, AIFS, Stormer, Fengwu, and Fuxi. Among them, NowcastNet integrates the differential equations with the deep learning model, Graph-EFM and AIFS both take the Earth's topology into consideration, while Stormer, Fengwu and Fuxi do not utilize physics and topology information. However, NowcastNet exclusively focuses on regional precipitation nowcasting, while PASSAT focuses on global, multi-variable and medium-term weather prediction. Graph-EFM and AIFS adopt hierarchical grid-mesh graph structures, similar to GraphCast. Stormer, Fengwu and Fuxi utilize attention-based structures similar to Pangu, and their focus is on improving long-term predictions via enhancing the training and inference strategies.

\noindent \textbf{Baseline NWP models.} PASSAT is also compared with the following operational NWP models: IFS T42 and IFS T63 \cite{rasp2020weatherbench}. IFS T42 and IFS T63 are the Integrated Forecast System (IFS) model run at two different resolutions, 2.8$^\circ$ and 1.9$^\circ$ respectively. We can observe that they are both finer than the 5.625$^\circ$ resolution of PASSAT, at the cost of being  computationally demanding in solving large systems of differential equations.

\noindent \textbf{Results.} As demonstrated in Figure \ref{figure_rmse}, PASSAT outperforms the other deep learning models in all weather variables across different lead times. The closest with PASSAT is GraphCast, which takes the topology of the Earth’s surface into consideration. However, GraphCast ignores the physics of the weather evolution, and thus has to use a more complex graph structure than PASSAT (twice in terms of the number of nodes and three times in terms of the number of edges).
ClimODE, despite of its physics-assisted structure, does not perform well. This phenomenon could be attributed to the following reasons: (i) ClimODE characterizes the evolution of the weather variables with the continuity equation, without considering the Earth-atmosphere interaction; (ii) ClimODE updates the velocity fields with a neural network, other than the Navier-Stokes equation; (iii) ClimODE ignores the topology information, and thus suffers from the distortions. In contrast, PASSAT benefits from both the physics information and the topology information, allowing it to achieve remarkably better performance.

\begin{figure}[htbp]
    \centering
    \includegraphics[width=40pc]{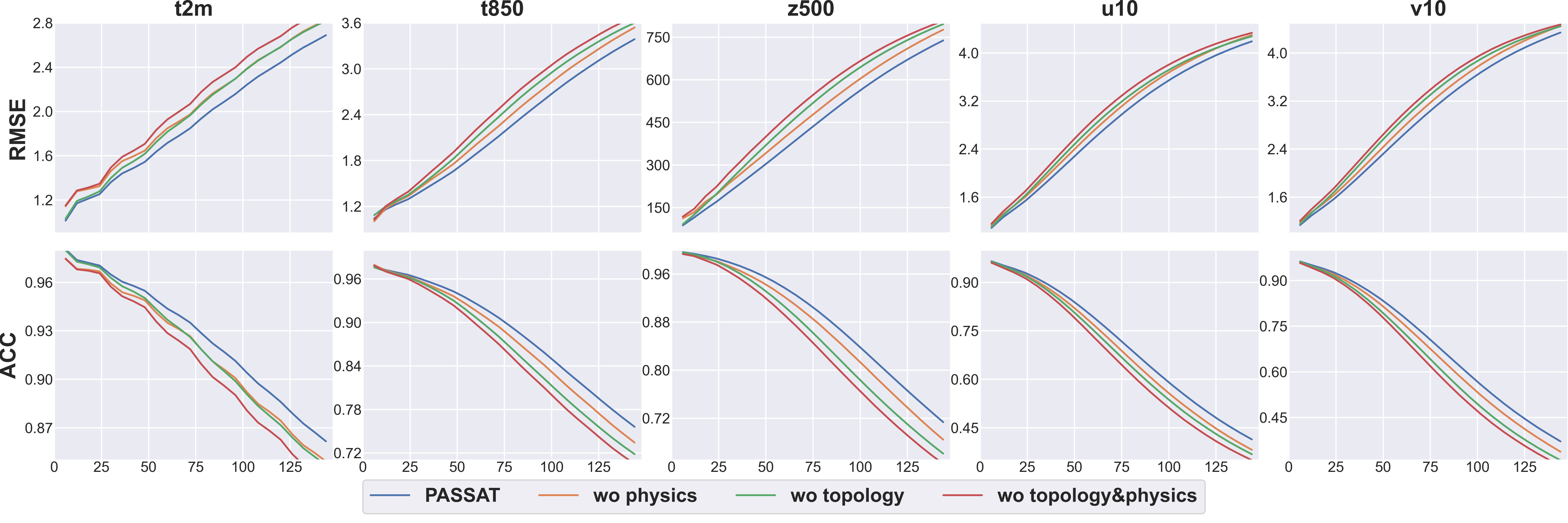}
	\caption{\textbf{Comparison between PASSAT and the three variants.} The x-axis represents the lead time in hours. Smaller RMSE and larger ACC values indicate better performance.}
\label{figure_ablation_rmse}
\end{figure}

The RMSEs and ACCs of IFS T42 and IFS T63 are from \cite{rasp2020weatherbench}, only including t2m, t850 and z500 for the lead times of 72 and 120 hours. Observe that PASSAT outperforms IFS T42, a pure physical model solved at a finer resolution (2.8$^\circ$). Improving the resolution of the physical model from 2.8$^\circ$ to 1.9$^\circ$, IFS T63 surpasses PASSAT and the other deep learning models, nevertheless at the cost of high computational complexity.
\section{Ablation Studies}

We conduct ablation studies to evaluate the effectiveness of the physics and topology information used in PASSAT. We compare with three models: (i) PASSAT (without topology), which constructs the graph neural network on the latitude-longitude planar grid, other than on the spherical manifold; (ii) PASSAT (without physics), which uses the spherical graph neural network to predict the weather in an end-to-end manner; (iii) PASSAT (without topology and physics), which uses the planar graph neural network to predict the weather in an end-to-end manner. 

The results are depicted in Figure \ref{figure_ablation_rmse}. PASSAT significantly outperforms the three variants across all weather variables, highlighting the importance of incorporating both physics and topology information. An interesting observation from our ablation studies is that different variables benefit from different sources of information. For t2m, the performance gains of respectively using the topology information and the physics information are almost the same. On the other hand, the rest variables of t850, z500, u10, and v10 benefit more from the topology information than from the physics information.

We do not evaluate the individual effects of the advection equation and the Naiver-Stokes equation, the two parts of the physics information. First, without the advection equation, it is unnecessary to update the velocity fields with the Naiver-Stokes equation. Second, estimating the future velocity fields with the spherical graph neural network, other than the Naiver-Stokes equation, shall require frequent calls of the velocity branch, leading to excessive back propagation and thus unaffordable memory consumption during training.
\section{Conclusions and Future Works}

In this paper, we propose PASSAT, a novel physics-assisted and topology-informed deep learning model for weather prediction. PASSAT seamlessly integrates the advection equation and the Navier-Stokes equation that govern the evolution of the weather variables and their velocity fields, with a graph neural network that estimates the complex Earth-atmosphere interaction and the initial velocity fields. PASSAT also takes the topology of the Earth's surface into consideration, during solving the equations and training the graph neural network. In the $5.625^\circ$-resolution ERA5 data set, PASSAT outperforms both the state-of-the-art deep learning-based weather prediction models and the operational numerical weather prediction model IFS T42. Our ablation studies demonstrate that both the physics information and the topology information are essential to the performance gain.

As future works, we will extend PASSAT in the following aspects: (i) Enhance PASSAT by incorporating more weather variables; (ii) Refine PASSAT through training over a data set with a finer resolution; (iii) Incorporate new time integration scheme that is more efficient than Euler's method, during both training and prediction. We expect that PASSAT is able to motivate more research efforts in combining physics, topology and historical observations for weather prediction.

\section{Ethical Statement}

There are no ethical issues.

\clearpage
\printbibliography

\appendix

\clearpage

\clearpage
\setcounter{page}{1}
%
%

\section{Data}
\subsection{Data Set}

We adopt the European Centre for Medium-Range Weather Forecasts Reanalysis V5 (ERA5) $5.625^\circ$-resolution data set from 1979 to 2018, provided by WeatherBench. Table \ref{tab4} summarizes the weather variables in the data set.
For more details, readers are referred to \href{https://confluence.ecmwf.int/display/CKB/ERA5\%3A+data+documentation}{ERA5}.

\begin{table*}[htb]
\caption{\textbf{Weather variables in the data set.}}
\centering
\resizebox{14cm}{!}{
\begin{tabular}{ccccc}
\toprule
\textbf{\textbf{Long name}} & \textbf{Short name} & \textbf{Description} & \textbf{Unit} & \textbf{Levels} \\
\midrule
geopotential & z500 & Proportional to the height of a pressure level & m$^2$s$^{-2}$ & 500hPa   \\
\midrule
temperature & t850 & Temperature & K & 850hPa   \\
\midrule
2m\_temperature & t2m & Temperature at 2m height above surface & K & -   \\
\midrule
10m\_u\_component\_of\_wind & u10 & Wind in longitude-direction at 10m height & ms$^{-1}$ & -   \\
\midrule
10m\_v\_component\_of\_wind & v10 & Wind in latitude-direction at 10m height & ms$^{-1}$ & -   \\
\midrule
land\_binary\_mask & lsm & Land-sea binary mask & (0/1) & -   \\
\midrule
orography & orography & Height of surface & m & -   \\

\bottomrule
\end{tabular}}
	\label{tab4}
\end{table*}

\subsection{Data Preprocessing}

\noindent \textbf{Normalization.} As the weather variables have diverse magnitudes, we use their means and standard deviations in 2006 to normalize the entire data set.

\noindent \textbf{Mapping the latitude-longitude grid to the sphere.} In PASSAT, we need to project a latitude-longitude point $(\theta,\phi)$ onto the unit sphere, as:
\begin{equation}
\begin{aligned}
\mathbf{s}(\theta,\phi) =
\left[\begin{array}{c}
\cos\theta \cos \phi \\
\cos\theta \sin\phi \\
\sin \theta
\end{array}\right] \in \mathbf{S}.
\end{aligned}
\end{equation}

\section{PASSAT}

\subsection{Structure} \label{overview}
In Table \ref{tab_PASSAT}, we illustrate the structures of PASSAT and its variants. For PASSAT without physics, we no longer need the velocity branch, and are able to merge the interaction branch with the backbone.

\begin{table*}[b]
\caption{\textbf{The structure of PASSAT and its variants.} The number in the brackets is the output dimension. }
\centering
\resizebox{14cm}{!}{
\begin{tabular}{ccccc}
\toprule
& \textbf{PASSAT} &  \textbf{w/o physics} & \textbf{w/o topology} &  \textbf{w/o topology\&physics}   \\
\midrule
\textbf{Num Nodes} & 2048 & 2048 & 2048 & 2048  \\
\midrule
\textbf{Num Edges} & 9152 & 9152 & 4000 & 4000   \\
\midrule
\textbf{Input Embedding} &  $ 1 \times $ MLP (48) & $ 1 \times $ MLP (45) & $1 \times $ MLP (48) &  $ 1 \times $ MLP (45)   \\
\midrule
\textbf{Backbone} & $2 \times$ Basic Block (48) &  \makecell[c]{[$2 \times$ Basic Block (45), \\ $1 \times$ Basic Block (90)]} & $2 \times$ Basic Block (48) &  \makecell[c]{[$2 \times$ Basic Block (45), \\ $1 \times$ Basic Block (90)]} \\
\midrule
\textbf{Velocity Branch} & $1 \times$ Basic Block (48) & - &   $1 \times$ Basic Block (48) & -  \\
\midrule
\textbf{Physics Branch} & \makecell[c]{[$2 \times$ Basic Block (48), \\ $1 \times$ Basic Block (24)]} & - &  \makecell[c]{[$2 \times$ Basic Block (48), \\ $1 \times$ Basic Block (24)]} & -  \\
\midrule
\textbf{Parameters} & 1.15 (M) & 1.15 (M) & 1.15(M) & 1.15(M)    \\
\bottomrule
\end{tabular}}
	\label{tab_PASSAT}
\end{table*}

\noindent \textbf{Graph.}
The spherical graph neural network of PASSAT is denoted as $(\mathcal{N}, \mathcal{E}, A)$, where $\mathcal{N}$ is the node set, $\mathcal{E}$ is the edge set and $A$ is the adjacency matrix.

Each node of PASSAT is represented by a spatial coordinate. Corresponding to the $5.625^\circ$-resolution data set, there are $2048$ nodes, represented as:
\begin{align}
 (\mathbf{n}_{0}, \mathbf{n}_{1}, \cdots, \mathbf{n}_{63},  \cdots,  \mathbf{n}_{2047})  = (\mathbf{s}(\phi_{0}, \theta_{0}), \mathbf{s}(\phi_{1}, \theta_{0}), \cdots, \nonumber \mathbf{s}(\phi_{63}, \theta_{0}), \cdots, \mathbf{s}(\phi_{63}, \theta_{31})).
\end{align}

The $(i,j)$-th element of the original adjacency matrix $A$ is given by the Haversine formula and Gaussian kernel:
\begin{equation}
(A)_{ij} = \exp(- 200 \times \text{Haversine}(\mathbf{n}_{i}, \mathbf{n}_{j})^2).
\end{equation}
To improve the computational efficiency, we prune $A$ by setting its elements to zero if they are below a given threshold. In the pruned adjacency matrix, the sparsest row has $5$ non-zeros. We also re-normalize $A$ to avoid exploding/vanishing gradients, as:
\begin{equation}
A \Rightarrow D^{-\frac{1}{2}}AD^{-\frac{1}{2}},
\end{equation}
where $D$ is a diagonal degree matrix, with $(D)_{ii} = \sum_{j}(A)_{ij}$.

We will use $A$ for feature extraction. In addition, we define the edge set $\mathcal{E}$ by $A$. If the $(i,j)$-th element of $A$ is non-zero, we say that nodes $\mathbf{n}_i$ and $\mathbf{n}_j$ are neighbors, and $(\mathbf{n}_i,\mathbf{n}_j)\in \mathcal{E}$. 

\noindent \textbf{Initial States of Nodes and Edges.}
We utilize $\mathbf{u}$ to stack all weather variables. The states of all weather variables at initial time determine the initial states of nodes and edges. For node $\mathbf{n}_{i}$, its initial state is:
\begin{equation}
\quad h^{\text{node}}_{i} = \mathbf{u}^{0}(\mathbf{n}_{i}),
\end{equation}
where $\mathbf{u}^{0}(\mathbf{n}_{i})$ denote the weather variables of spatial coordinate $\mathbf{n}_{i}$ at step $0$, respectively. For edge $(\mathbf{n}_i,\mathbf{n}_j)$ connecting nodes $\mathbf{n}_{i}$ and $\mathbf{n}_{j}$, its initial state is:
\begin{equation}
h^{\text{edge}}_{ij} = (|\theta_{i}-\theta_{j}|, |\phi_{i}-\phi_{j}|, \text{Haversine}(\mathbf{n}_{i}, \mathbf{n}_{j})).
\end{equation}


\noindent \textbf{Basic Blocks in PASSAT.}
The basic block of PASSAT consists of a node-edge connection block and node-node aggregation block. In node-edge connection, the node to edge message passing (blue arrow) involves concatenating the hidden states of the two nodes $\mathbf{n}_i$ and $\mathbf{n}_j$ connected by each edge $(\mathbf{n}_i, \mathbf{n}_j)$, written as:
\begin{equation}\label{edgecon}
h^{\text{edge}}_{ij} \Rightarrow (h^{\text{edge}}_{ij}, h^{\text{node}}_{i}, h^{\text{node}}_{j}),
\end{equation}
where $h^{\text{edge}}_{ij}$ is the hidden state of edge $(\mathbf{n}_i, \mathbf{n}_j)$, while $ h^{\text{node}}_{i}$ and  $h^{\text{node}}_{j}$ are the hidden states of nodes $\mathbf{n}_i$ and $\mathbf{n}_j$, respectively.
The edge to node message passing (orange arrow) is to concatenate the sum of the hidden states of edges that each node $\mathbf{n}_i$ connects, written as:
\begin{equation}\label{nodecon}
h^{\text{node}}_{i} \Rightarrow ( h^{\text{node}}_{i}, \sum_{j:(\mathbf{n}_i, \mathbf{n}_j) \in \mathcal{E} }h^{\text{edge}}_{ij}).
\end{equation}

The node-node aggregation uses the adjacency matrix $A$ to aggregate the hidden states of all nodes (green arrow), and concatenate the results to update the hidden states of all no- des, written as:
\begin{equation}\label{graphconv}
h^{\text{node}}_{i} \Rightarrow ( h^{\text{node}}_{i}, \sum_{j}(A)_{ij}h^{\text{node}}_{j}).
\end{equation}

\section{Performance Metrics}

\noindent \textbf{Root mean square error (RMSE).} Given a weather variable $u$, we suppose that the initial time is $0$ and that the lead time is $\tau$. The RMSE is defined as:
\begin{equation}
\text{RMSE}(\tau) =  \sqrt{\frac{1}{|\mathbf{S}_{\text{d}}|} \sum_{\mathbf{s} \in \mathbf{S}_{\text{d}}}a_{\mathbf{s}} (u(\tau,\mathbf{s})-\check{u}(\tau,\mathbf{s}))^{2}}.
\end{equation}
Therein, $\mathbf{S}_{\text{d}} \subset \mathbf{S}$ is the set of discrete spatial coordinates, while $u(\tau,\mathbf{s})$ and $\check{u}(\tau,\mathbf{s})$ are the prediction and observation of $u$ at lead time $\tau$ and spatial coordinate $\mathbf{s}$. The weight $a(\mathbf{s})$ is defined as:
\begin{equation}
a(\mathbf{s}) = \frac{\cos \theta}{\frac{1}{|\mathbf{S}_{\text{d}}|} \sum_{\mathbf{s'} \in \mathbf{S}_{\text{d}}} \cos \theta'}.
\end{equation}
The reported RMSE is the average over all initial times.

\noindent \textbf{Anomaly correlation coefficient (ACC).} Given a weather variable $u$, we suppose that the initial time is $0$ and that the lead time is $\tau$. The ACC is defined as:
\begin{align}
\text{ACC}(\tau) =  \frac{\sum_{\mathbf{s} \in \mathbf{S}_{\text{d}}}a_{\mathbf{s}} \text{Clim}({u}(\tau,\mathbf{s})) \text{Clim}(\check{u}(\tau,\mathbf{s}))}
{\sqrt{\sum_{\mathbf{s} \in \mathbf{S}_{\text{d}}}a_{\mathbf{s}} \text{Clim}({u}(\tau,\mathbf{s}))^2 \times \sum_{\mathbf{s} \in \mathbf{S}_{\text{d}}}a_{\mathbf{s}}\text{Clim}(\check{u}(\tau,\mathbf{s}))^2}}. \label{acc}
\end{align}
Therein, we define:
\begin{align}
\text{Clim}(u(\tau,\mathbf{s})) = u(\tau,\mathbf{s}) - C(\mathbf{s}) - \frac{1}{|\mathbf{S}_{\text{d}}|}\sum_{\mathbf{s}' \in \mathbf{S}_{\text{d}}}(u(\tau,\mathbf{s}') - C(\mathbf{s}')),
\end{align}
where $C(\mathbf{s})$ is the climatological mean of weather variable $u$ at spatial coordinate $\mathbf{s}$, computed using the ERA5 data set of 2006. The reported ACC is the average over all initial times.

\section{Training Details}

\subsection{Loss Functions}

Given a weather variable $u$ and at any initial time denoted by $0$, the loss function of ClimODE, GraphCast, Pangu, FourCastNet, and SFNO is given by:
\begin{equation}
\mathcal{L}_{\text{basic}} = \frac{1}{T_a}\sum_{\tau \in 1:T_a} \frac{1}{|\mathbf{S}_{\text{d}}|} \sum_{\mathbf{s} \in \mathbf{S}_{\text{d}}} (u(\tau,\mathbf{s}) - \check{u}(\tau,\mathbf{s}))^{2}.
\end{equation}
Therein, $\mathbf{S}_{\text{d}} \subset \mathbf{S}$ is the set of discrete spatial coordinates, while $u(\tau,\mathbf{s})$ and $\check{u}(\tau,\mathbf{s})$ are the prediction and observation of $u$ at lead time $\tau$ and spatial coordinate $\mathbf{s}$. We use $T_a$ to denote the number of autoregressive steps. Then, $\mathcal{L}_{\text{basic}}$ is averaged over all weather variables and all initial times.

The predictions of PASSAT also involve the initial motion fields, whose values must be controlled. Therefore, we introduce several penalty terms to the loss function. Given a motion field $v$ and at any initial time denoted by $0$, we define:
\begin{equation}
\mathcal{L}_{\text{velocity}} = \mathcal{L}_{\text{velocity}}^{1}+\mathcal{L}_{\text{velocity}}^{2}+\mathcal{L}_{\text{velocity}}^{3} ,
\end{equation}
where
\begin{equation}
\mathcal{L}_{\text{velocity}}^{1} = \frac{\lambda_1}{2|\mathbf{S}_{\text{d}}|} \sum_{\mathbf{s} \in \mathbf{S}_{\text{d}}} [(v_{\theta}(0,\mathbf{s}))^{2} + (v_{\phi}(0,\mathbf{s}))^{2}],
\end{equation}
\begin{equation}
\mathcal{L}_{\text{velocity}}^{2} = \frac{\lambda_2}{2|\mathbf{S}_{\text{d}}|} \sum_{\mathbf{s} \in \mathbf{S}_{\text{d}}} [(\frac{\partial }{\partial \theta}v_{\theta}(0,\mathbf{s}))^{2} + (\frac{\partial }{\partial \theta} v_{\phi}(0,\mathbf{s}))^{2}],
\end{equation}
\begin{equation}
\hspace{-0.2em} \mathcal{L}_{\text{velocity}}^{3} = \frac{\lambda_3}{2|\mathbf{S}_{\text{d}}|} \sum_{\mathbf{s} \in \mathbf{S}_{\text{d}}} [(\frac{\partial }{\partial \phi}v_{\theta}(0,\mathbf{s}))^{2} + (\frac{\partial }{\partial \phi}v_{\phi}(0,\mathbf{s}))^{2}].
\end{equation}
Therein, $\lambda_1=10$, $\lambda_2=1$ and $\lambda_3=1$ are constants to penalize the initial motion field and its smoothness. Then, $\mathcal{L}_{\text{velocity}}$ is averaged over all velocity fields and all initial times.



In summary, the loss function of PASSAT is given by $\mathcal{L} = \mathcal{L}_{\text{basic}} + \mathcal{L}_{\text{velocity}}$.

\subsection{Training Strategies}

We train PASSAT and the baseline deep learning models from scratch in an autoregressive manner; that is, we treat the current predictions as observations and feed them back to the models to generate future predictions. We train each model to predict the weather variables for the future 6, 12, 18 and 24 hours.

%
%

We use the AdamW optimizer with parameters $\beta_{1} = 0.9$ and $\beta_{2} = 0.999$, and set the weight decay as 0.05. Gradient clipping is also employed, with a maximum norm value of 1. We use PyTorch for training, validation and prediction, on four GeForece RTX 2080. The batch size is 2 per GPU (8 in total). we adjust the learning rate with the Cosine-LR-Scheduler, and set the maximum and minimum learning rates to 1e-3 and 3e-7, respectively. The number of epochs is 50.

\section{Baseline Deep Learning Models}

We consider the following baseline deep learning models: GraphCast, Pangu, ClimODE, FourCastNet, and SFNO. For fair comparisons, we unify the number of parameters to the same magnitude (around 1.15 million). We show their modified structures in Tables \ref{tab6}--\ref{tab10}.

\begin{table}[htbp]
\caption{\textbf{The modification of GraphCast.}}
\centering
\begin{tabular}{ccc}
\toprule
 & \textbf{Original} & \textbf{Modified (this paper)}  \\
\midrule
\textbf{Num Nodes} & 1079202 & 4610  \\
\midrule
\textbf{Num Edges} & 5061126 & 29732   \\
\midrule
\textbf{Mesh Level} & 6 & 4   \\
\midrule
\textbf{ \makecell[c]{Embedding \\ Dimension}} & 512 & 151  \\
\midrule
\textbf{Processer Layers} & 16 & 4   \\
\midrule
\textbf{Parameters} & 36.7 (M) & 1.15 (M)    \\
\bottomrule
\end{tabular}
	\label{tab6}
\end{table}

\begin{table}[htbp]
\caption{\textbf{The modification of Pangu.}}
\centering
\begin{tabular}{ccc}
\toprule
 & \textbf{Original} & \textbf{Modified (this paper)}  \\
\midrule
\textbf{ \makecell[c]{Embedding \\ Dimension}} & 192 & 24  \\
\midrule
\textbf{Num Layers} & \makecell[c]{[2(192), 6(384), \\6(384), 2(192)]} & \makecell[c]{[2(24), 6(48),\\ 6(48), 2(24)]}   \\
\midrule
\textbf{Num Heads} & [6, 12, 12, 6] & [2, 4, 4, 2]    \\
\midrule
\textbf{Patch Sizes} & (2, 4, 4) & (1, 1)    \\
\midrule
\textbf{Window Sizes} & (2, 6, 12) & (4, 8)    \\
\midrule
\textbf{Parameters} & 64 (M) & 1.20 (M)    \\
\bottomrule
\end{tabular}
	\label{tab7}
\end{table}

\begin{table}[htbp]
\caption{\textbf{The modification of ClimODE.}}
\centering
\begin{tabular}{ccc}
\toprule
 & \textbf{Original} & \textbf{Modified (this paper)}  \\
\midrule
\textbf{\makecell[c]{Num Layers \\ (Velocity)}} & [5(128), 3(64), 2(10)] & [5(64), 3(32), 2(10))]   \\
\midrule
\textbf{\makecell[c]{Num Layers \\ (Noise)}} & [3(128), 2(64), 2(10)] & [4(96), 3(72), 2(5)]    \\
\midrule
\textbf{Parameters} & 2.8 (M) & 1.33 (M)    \\
\bottomrule
\end{tabular}
	\label{tab8}
\end{table}

\begin{table}[htbp]
\caption{\textbf{The modification of FourCastNet.}}
\centering
\begin{tabular}{ccc}
\toprule
 & \textbf{Original} & \textbf{Modified (this paper)}  \\
\midrule
\textbf{Embedding Dimension} & 768 & 172   \\
\midrule
\textbf{Num Layers} & 12 & 4    \\
\midrule
\textbf{Num Blocks} & 16 & 4    \\
\midrule
\textbf{Pacth Size} & (16, 16) & (2, 2)    \\
\midrule
\textbf{Parameters} & 59.1 (M) & 1.17 (M)    \\
\bottomrule
\end{tabular}
	\label{tab9}
\end{table}

\begin{table}[htbp]
\caption{\textbf{The modification of SFNO.}}
\centering
\begin{tabular}{ccc}
\toprule
 & \textbf{Original} & \textbf{Modified (this paper)}  \\
\midrule
\textbf{Embedding Dimension} & 1024 & 65   \\
\midrule
\textbf{Scaler factor} & 3.0 & 1.0   \\
\midrule
\textbf{Rank} & 128 & 1   \\
\midrule
\textbf{Parameters} & 107 (M) & 1.15 (M)    \\
\bottomrule
\end{tabular}
	\label{tab10}
\end{table}

\section{Comparisons with Baseline Models} \label{baseline-more}

We compare PASSAT with the other models in Tables \ref{comp_rmse} and \ref{comp_acc} in terms of RMSE and ACC. PASSAT is the best among all deep learning models. Graphcast and FourCastNet are the closest two. IFS T63 outperforms IFS T42 and all deep learning models, but is computationally expensive.

\begin{table*}[tb]
\caption{\textbf{Comparisons between PASSAT and the other models in terms of RMSEs, over the test
set.} For each lead time and each weather variable, the best model RMSE is in \textbf{bold} and the second best model RMSE is \underline{underlined}.}
\centering
\resizebox{14cm}{!}{
\begin{tabular}{cccccccccc}
\toprule
\textbf{} & \textbf{Lead Time (h)} & \textbf{PASSAT} & \textbf{GraphCast} & \textbf{ClimODE} & \textbf{Pangu} & \textbf{FourCastNet} & \textbf{SFNO} & \textbf{IFS T42} & \textbf{IFS T63} \\
\midrule
\textbf{Physics-assisted} & - & Yes & No & Yes & No & No & No & - & -  \\
\midrule
\textbf{Topology-informed} & - & Yes &  Yes & No & No & No & Yes & - & -  \\
\midrule
\textbf{Parameters} (M) & - & 1.15 & 1.15 & 1.33 & 1.20 & 1.17 & 1.15 & - & -  \\
\midrule
 & 24 & \underline{1.25} & \textbf{1.21} & 1.59 & 1.31 & 1.29 & 1.43 & - & -  \\
 & 48 & \underline{1.54} & \textbf{1.52} & 2.23 & 1.63 & 1.58 & 1.75 & - & -  \\
\textbf{t2m} & 72 & \underline{1.85} & \textbf{1.84} & NAN & 1.94 & 1.87 & 2.02 & 3.21 & 2.04  \\
 & 96 & \textbf{2.16} & \underline{2.17} & NAN & 2.26 & \textbf{2.16} & 2.30 & - & -  \\
 & 120 & \textbf{2.44} & \underline{2.48} & NAN & 2.55 & \textbf{2.44} & 2.55 & 3.69 & \textbf{2.44}  \\
 & 144 & \underline{2.69} & 2.75 & NAN & 2.80 & \textbf{2.68} & 2.76 & - & -  \\
\midrule
 & 24 & \textbf{1.29} & \textbf{1.29} & 1.46 & 1.40 & \underline{1.39} & 1.51 & - & -  \\
 & 48 & \textbf{1.66} & \underline{1.69} & 2.26 & 1.83 & 1.78 & 1.91 & - & -  \\
\textbf{t850} & 72 & \underline{2.12} & 2.17 & NAN & 2.31 & 2.24 & 2.37 & 3.09 & \textbf{1.85}  \\
 & 96 & \textbf{2.59} & \underline{2.67} & NAN & 2.79 & 2.71 & 2.84 & - & -  \\
 & 120 & \underline{3.03} & 3.12 & NAN & 3.21 & 3.12 & 3.25 & 3.83 & \textbf{2.52}  \\
 & 144 & \textbf{3.39} & 3.49 & NAN & 3.55 & \underline{3.47} & 3.59 & - & -  \\
\midrule
 & 24 & \textbf{171} & \underline{180} & 209 & 196 & 188 & 205 & - & -  \\
 & 48 & \textbf{293} & \underline{307} & 452 & 337 & 313 & 342 & - & -  \\
\textbf{z500} & 72 & \underline{420} & 438 & NAN & 477 & 442 & 478 & 489 & \textbf{268}  \\
 & 96 & \textbf{543} & \underline{565} & NAN & 605 & 566 & 607 & - & -  \\
 & 120 & \underline{651} & 675 & NAN & 712 & 674 & 718 & 743 & \textbf{463}  \\
 & 144 & \textbf{738} & 767 & NAN & 797 & \underline{760} & 807 & - & -  \\
\midrule
 & 24 & \textbf{1.54} & \underline{1.59} & 1.77 & 1.68 & 1.68 & 1.83 & - & -  \\
 & 48 & \textbf{2.22} & \underline{2.29} & 2.73 & 2.41 & 2.38 & 2.50 & - & -  \\
\textbf{u10} & 72 & \textbf{2.90} & \underline{2.97} & NAN & 3.08 & 3.05 & 3.18 & - & -  \\
 & 96 & \textbf{3.46} & \underline{3.54} & NAN & 3.61 & 3.60 & 3.73 & - & -  \\
 & 120 & \textbf{3.89} & \underline{3.97} & NAN & 4.01 & 4.00 & 4.13 & - & -  \\
 & 144 & \textbf{4.19} & \underline{4.27} & NAN  & 4.29 & 4.29 & 4.41 & - & -  \\
\midrule
 & 24 & \textbf{1.57} & \underline{1.62} & 1.80 & 1.71 & 1.72 & 1.86 & - & -  \\
 & 48 & \textbf{2.26} & \underline{2.32} & 2.73 & 2.44 & 2.42 & 2.53 & - & -  \\
\textbf{v10} & 72 & \textbf{2.95} & \underline{3.03} & NAN & 3.14 & 3.12 & 3.24 & - & -  \\
 & 96 & \textbf{3.55} & \underline{3.64} & NAN & 3.70 & 3.70 & 3.83 & - & -  \\
 & 120 & \textbf{4.01} & \underline{4.10} & NAN & 4.13 & 4.14 & 4.27 & - & -  \\
 & 144 & \textbf{4.34} & 4.44 & NAN & \underline{4.43} & 4.44 & 4.57 & - & -  \\
\bottomrule
\end{tabular}}
	\label{comp_rmse}
\end{table*}

\begin{table*}[tb]
\caption{\textbf{Comparisons between PASSAT and the other models in terms of ACCs, over the test
set.} For each lead time and each weather variable, the best model ACC is in \textbf{bold} and the second best model ACC is \underline{underlined}.}
\centering
\resizebox{14cm}{!}{
\begin{tabular}{cccccccccc}
\toprule
\textbf{} & \textbf{Lead Time (h)} & \textbf{PASSAT} & \textbf{GraphCast} & \textbf{ClimODE} & \textbf{Pangu} & \textbf{FourCastNet} & \textbf{SFNO} & \textbf{IFS T42} & \textbf{IFS T63} \\
\midrule
\textbf{Physics-assisted} & - & Yes & No & Yes & No & No & No & - & -  \\
\midrule
\textbf{Topology-informed} & - & Yes &  Yes & No & No & No & Yes & - & -  \\
\midrule
\textbf{Parameters} (M) & - & 1.15 & 1.15 & 1.33 & 1.20 & 1.17 & 1.15 & - & -  \\
\midrule
 & 24 & \underline{0.970} & \textbf{0.972} & 0.950 & 0.968 & 0.969 & 0.962 & - & -  \\
 & 48 & \underline{0.955} & \textbf{0.956} & 0.910 & 0.950 & 0.953 & 0.942 & - & -  \\
\textbf{t2m} & 72 & 0.935 & \underline{0.937} & 0.510 & 0.929 & 0.934 & 0.922 & 0.870 & \textbf{0.940}  \\
 & 96 & \underline{0.911} & \textbf{0.912} & 0.020 & 0.903 & \underline{0.911} & 0.899 & - & -  \\
 & 120 & 0.886 & 0.886 & NAN & 0.877 & \underline{0.888} & 0.875 & 0.830 & \textbf{0.920}  \\
 & 144 & \underline{0.862} & 0.861 & NAN  & 0.853 & \textbf{0.865} & 0.853 & - & -  \\
\midrule
 & 24 & \textbf{0.966} & \textbf{0.966} & \underline{0.960} & \underline{0.960} & \underline{0.960} & 0.953 & - & -  \\
 & 48 & \textbf{0.943} & \underline{0.941} & 0.900 & 0.931 & 0.934 & 0.924 & - & -  \\
\textbf{t850} & 72 & \underline{0.906} & 0.902 & 0.450 & 0.888 & 0.895 & 0.881 & 0.860 & \textbf{0.940}  \\
 & 96 & \textbf{0.859} & \underline{0.851} & 0.030 & 0.835 & 0.845 & 0.828 & - & -  \\
 & 120 & \underline{0.806} & 0.796 & NAN & 0.780 & 0.792 & 0.773 & 0.780 & \textbf{0.900}  \\
 & 144 & \textbf{0.756} & \underline{0.745} & NAN & 0.732 & 0.743 & 0.724 & - & -  \\
\midrule
 & 24 & \textbf{0.986} & \underline{0.984} & 0.980 & 0.981 & 0.983 & 0.979 & - & -  \\
 & 48 & \textbf{0.958} & \underline{0.954} & 0.910 & 0.944 & 0.951 & 0.942 & - & -  \\
\textbf{z500} & 72 & \underline{0.912} & 0.904 & 0.470 & 0.885 & 0.901 & 0.883 & 0.900 & \textbf{0.970}  \\
 & 96 & \textbf{0.849} & \underline{0.838} & NAN & 0.813 & 0.835 & 0.809 & - & -  \\
 & 120 & \underline{0.780} & 0.766 & NAN & 0.737 & 0.763 & 0.731 & \underline{0.780} & \textbf{0.910}  \\
 & 144 & \textbf{0.713} & \underline{0.698} & NAN & 0.668 & 0.695 & 0.658 & - & -  \\
\midrule
 & 24 & \textbf{0.929} & \underline{0.924} & 0.910 & 0.914 & 0.914 & 0.898 & - & -  \\
 & 48 & \textbf{0.848} & \underline{0.838} & 0.760 & 0.818 & 0.824 & 0.802 & - & -  \\
\textbf{u10} & 72 & \textbf{0.733} & \underline{0.718} & 0.300 & 0.692 & 0.702 & 0.672 & - & -  \\
 & 96 & \textbf{0.610} & \underline{0.593} & NAN & 0.565 & 0.578 & 0.541 & - & -  \\
 & 120 & \textbf{0.501} & \underline{0.485} & NAN & 0.457 & 0.472 & 0.434 & - & -  \\
 & 144 & \textbf{0.414} & \underline{0.399} & NAN & 0.373 & 0.389 & 0.349 & - & -  \\
\midrule
 & 24 & \textbf{0.926} & \underline{0.921} & 0.900 & 0.912 & 0.911 & 0.895 & - & -  \\
 & 48 & \textbf{0.842} & \underline{0.833} & 0.770 & 0.814 & 0.818 & 0.798 & - & -  \\
\textbf{v10} & 72 & \textbf{0.722} & \underline{0.708} & 0.320 & 0.680 & 0.689 & 0.662 & - & -  \\
 & 96 & \textbf{0.589} & \underline{0.572} & 0.010 & 0.542 & 0.553 & 0.519 & - & -  \\
 & 120 & \textbf{0.468} & \underline{0.452} & NAN & 0.420 & 0.435 & 0.399 & - & -  \\
 & 144 & \textbf{0.372} & \underline{0.355} & NAN & 0.324 & 0.344 & 0.307 & - & -  \\
\bottomrule
\end{tabular}}
	\label{comp_acc}
\end{table*}

\section{Ablation Studies} \label{ablation-more}
We also compare PASSAT with its variants in Tables \ref{abla_rmse} and \ref{abla_acc}, in terms of RMSE and ACC. It demonstrates that both the physics information and the topology information are critical to the performance gain.

\begin{table*}[tb]
\caption{\textbf{Ablation studies in terms of RMSE, over the test set.} For each lead time and each wea- ther variable, the best model RMSE is in \textbf{bold} and the second best model RMSE is \underline{underlined}.}
\centering
\resizebox{13cm}{!}{
\begin{tabular}{cccccc}
\toprule
\textbf{ } & \textbf{Lead Time (h)} & \textbf{PASSAT} & \textbf{w/o physics} & \textbf{w/o topology} & \textbf{w/o topology\&physics} \\
\midrule
\textbf{Physics-assisted} & - & Yes & No & Yes & No \\
\midrule
\textbf{Topology-informed} & - & Yes &  Yes & No & No \\
\midrule
\textbf{Parameters} (M) & - & 1.15 & 1.15 & 1.15 & 1.15 \\
\midrule
 & 24 & \textbf{1.25} & 1.33 & \underline{1.28} & 1.35 \\
 & 48 & \textbf{1.54} & 1.65 & \underline{1.62} & 1.71 \\
\textbf{t2m} & 72 & \textbf{1.85} & \underline{1.97} & \underline{1.97} & 2.07 \\
 & 96 & \textbf{2.16} & \underline{2.30} & \underline{2.30} & 2.40 \\
 & 120 & \textbf{2.44} & \underline{2.58} & \underline{2.58} & 2.68 \\
 & 144 & \textbf{2.69} & 2.83 & \underline{2.82} & 2.91 \\
\midrule
 & 24 & \textbf{1.29} & \underline{1.33} & 1.35 & 1.39 \\
 & 48 & \textbf{1.66} & \underline{1.76} & 1.82 & 1.91 \\
\textbf{t850} & 72 & \textbf{2.12} & \underline{2.25} & 2.37 & 2.47 \\
 & 96 & \textbf{2.59} & \underline{2.75} & 2.87 & 2.97 \\
 & 120 & \textbf{3.03} & \underline{3.18} & 3.28 & 3.37 \\
 & 144 & \textbf{3.39} & \underline{3.54} & 3.60 & 3.68 \\
\midrule
 & 24 & \textbf{171} & \underline{197} & 199 & 225 \\
 & 48 & \textbf{293} & \underline{331} & 357 & 387 \\
\textbf{z500} & 72 & \textbf{420} & \underline{462} & 502 & 531 \\
 & 96 & \textbf{543} & \underline{585} & 626 & 649 \\
 & 120 & \textbf{651} & \underline{690} & 723 & 740 \\
 & 144 & \textbf{738} & \underline{776} & 796 & 807 \\
\midrule
 & 24 & \textbf{1.54} & \underline{1.61} & 1.63 & 1.70 \\
 & 48 & \textbf{2.22} & \underline{2.34} & 2.42 & 2.51 \\
\textbf{u10} & 72 & \textbf{2.90} & \underline{3.03} & 3.12 & 3.21 \\
 & 96 & \textbf{3.46} & \underline{3.60} & 3.65 & 3.74 \\
 & 120 & \textbf{3.89} & \underline{4.01} & 4.02 & 4.10 \\
 & 144 & \textbf{4.19} & 4.30 & \underline{4.28} & 4.34 \\
\midrule
 & 24 & \textbf{1.57} & \underline{1.64} & 1.69 & 1.75 \\
 & 48 & \textbf{2.26} & \underline{2.37} & 2.50 & 2.58 \\
\textbf{v10} & 72 & \textbf{2.95} & \underline{3.09} & 3.22 & 3.31 \\
 & 96 & \textbf{3.55} & \underline{3.69} & 3.79 & 3.86 \\
 & 120 & \textbf{4.01} & \underline{4.13} & 4.18 & 4.23 \\
 & 144 & \textbf{4.34} & 4.45 & \underline{4.44} & 4.47 \\
\bottomrule
\end{tabular}}
	\label{abla_rmse}
\end{table*}

\begin{table*}[tb]
\caption{\textbf{Ablation studies in terms of ACC, over the test set.} For each lead time and each wea- ther variable, the best model ACC is in \textbf{bold} and the second best model ACC is \underline{underlined}.}
\centering
\resizebox{13cm}{!}{
\begin{tabular}{cccccc}
\toprule
\textbf{ } & \textbf{Lead Time (h)} & \textbf{PASSAT} & \textbf{w/o physics} & \textbf{w/o topology} & \textbf{w/o topology\&physics} \\
\midrule
\textbf{Physics-assisted} & - & Yes & No & Yes & No \\
\midrule
\textbf{Topology-informed} & - & Yes &  Yes & No & No \\
\midrule
\textbf{Parameters} (M) & - & 1.15 & 1.15 & 1.15 & 1.15 \\
\midrule
 & 24 & \textbf{0.970} & 0.967 & \underline{0.969} & 0.966 \\
 & 48 & \textbf{0.955} & 0.949 & \underline{0.950} & 0.945 \\
\textbf{t2m} & 72 & \textbf{0.935} & \underline{0.927} & 0.926 & 0.919 \\
 & 96 & \textbf{0.911} & \underline{0.901} & 0.899 & 0.890 \\
 & 120 & \textbf{0.886} & \underline{0.874} & 0.872 & 0.863 \\
 & 144 & \textbf{0.862} & \underline{0.849} & 0.847 & 0.838 \\
\midrule
 & 24 & \textbf{0.966} & \underline{0.964} & 0.962 & 0.960 \\
 & 48 & \textbf{0.943} & \underline{0.936} & 0.930 & 0.924 \\
\textbf{t850} & 72 & \textbf{0.906} & \underline{0.894} & 0.881 & 0.870 \\
 & 96 & \textbf{0.859} & \underline{0.842} & 0.823 & 0.810 \\
 & 120 & \textbf{0.806} & \underline{0.786} & 0.767 & 0.754 \\
 & 144 & \textbf{0.756} & \underline{0.734} & 0.718 & 0.705 \\
\midrule
 & 24 & \textbf{0.986} & \underline{0.981} & 0.980 & 0.975 \\
 & 48 & \textbf{0.958} & \underline{0.946} & 0.936 & 0.925 \\
\textbf{z500} & 72 & \textbf{0.912} & \underline{0.893} & 0.871 & 0.855 \\
 & 96 & \textbf{0.849} & \underline{0.825} & 0.796 & 0.778 \\
 & 120 & \textbf{0.780} & \underline{0.752} & 0.724 & 0.705 \\
 & 144 & \textbf{0.713} & \underline{0.684} & 0.661 & 0.644 \\
\midrule
 & 24 & \textbf{0.929} & \underline{0.922} & 0.920 & 0.913 \\
 & 48 & \textbf{0.848} & \underline{0.830} & 0.817 & 0.802 \\
\textbf{u10} & 72 & \textbf{0.733} & \underline{0.705} & 0.685 & 0.662 \\
 & 96 & \textbf{0.610} & \underline{0.577} & 0.557 & 0.532 \\
 & 120 & \textbf{0.501} & \underline{0.468} & 0.450 & 0.427 \\
 & 144 & \textbf{0.414} & \underline{0.382} & 0.369 & 0.350 \\
\midrule
 & 24 & \textbf{0.926} & \underline{0.920} & 0.915 & 0.907 \\
 & 48 & \textbf{0.842} & \underline{0.825} & 0.804 & 0.790 \\
\textbf{v10} & 72 & \textbf{0.722} & \underline{0.693} & 0.661 & 0.638 \\
 & 96 & \textbf{0.589} & \underline{0.555} & 0.516 & 0.493 \\
 & 120 & \textbf{0.468} & \underline{0.435} & 0.398 & 0.378 \\
 & 144 & \textbf{0.372} & \underline{0.338} & 0.310 & 0.293 \\
\bottomrule
\end{tabular}}
	\label{abla_acc}
\end{table*}

\section{Visualizations}

\begin{figure*}[tb]
    \centering
    \includegraphics[width=38pc]{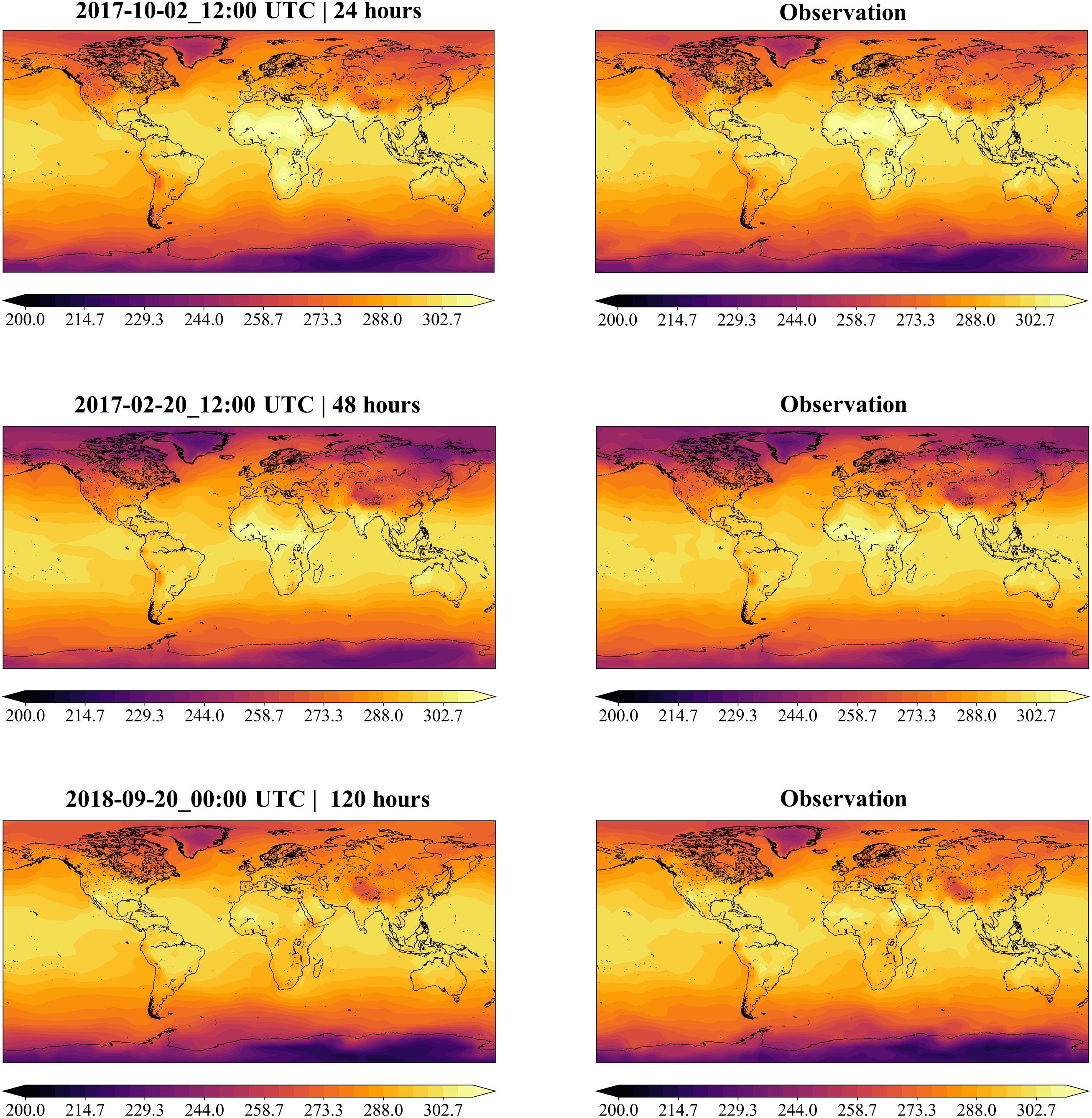}
	\caption{\textbf{Prediction visualization of t2m.}}
\label{visual_t2m}
\end{figure*}

\begin{figure*}[tb]
    \centering
    \includegraphics[width=38pc]{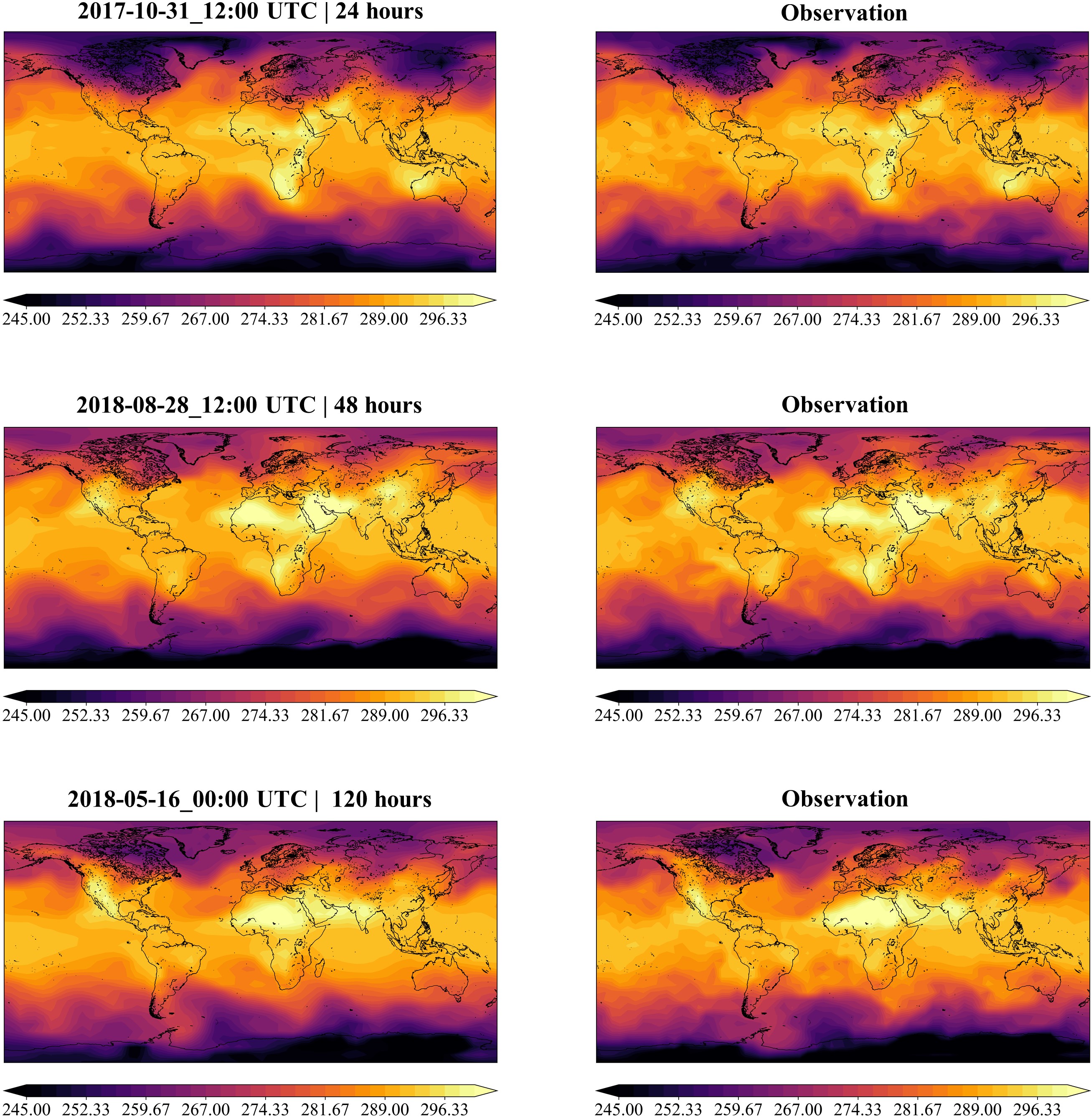}
	\caption{\textbf{Prediction visualization of t.}}
\label{visual_t850}
\end{figure*}

\begin{figure*}[tb]
    \centering
    \includegraphics[width=38pc]{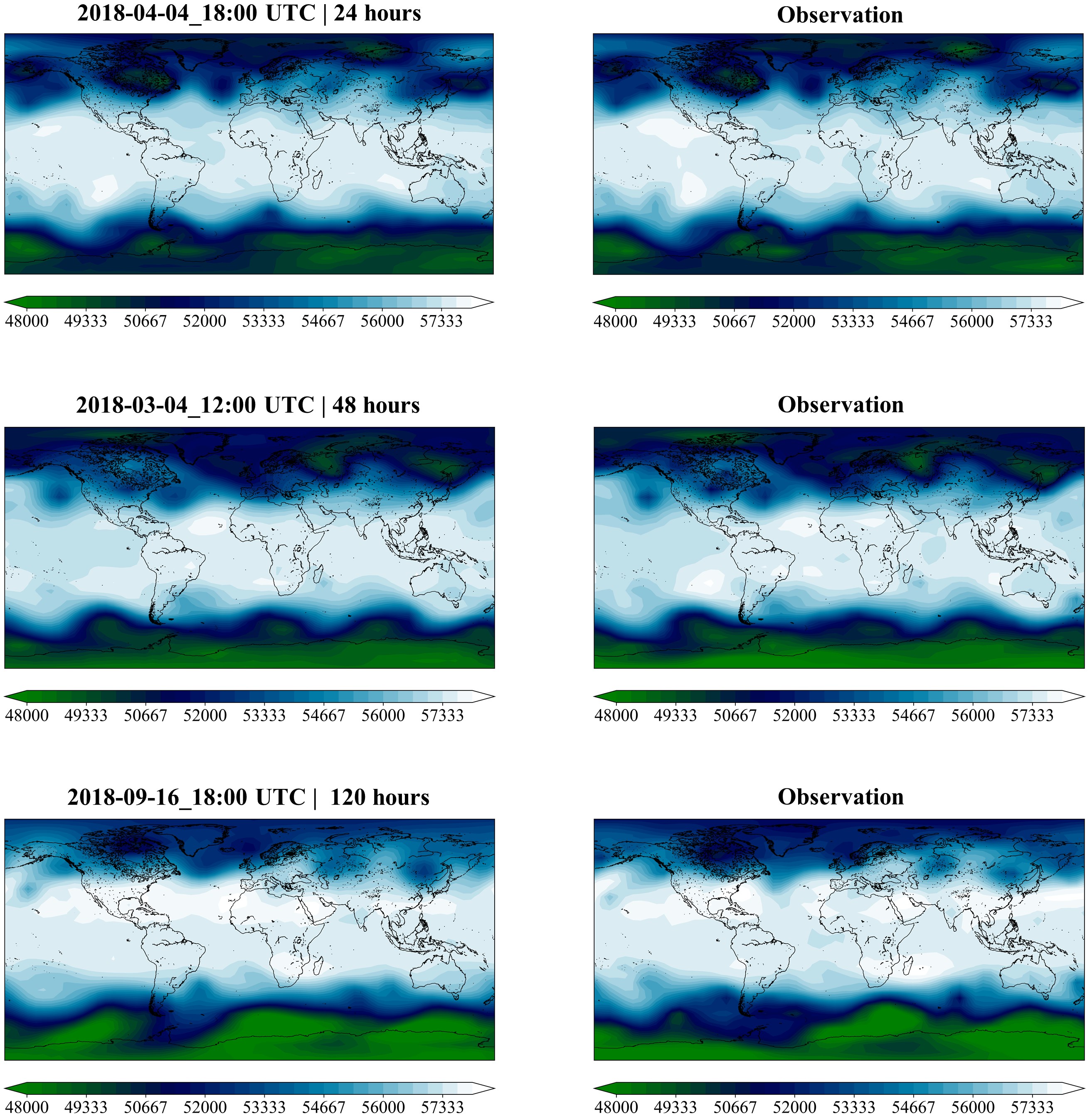}
	\caption{\textbf{Prediction visualization of z.}}
\label{visual_z500}
\end{figure*}

\begin{figure*}[tb]
    \centering
    \includegraphics[width=38pc]{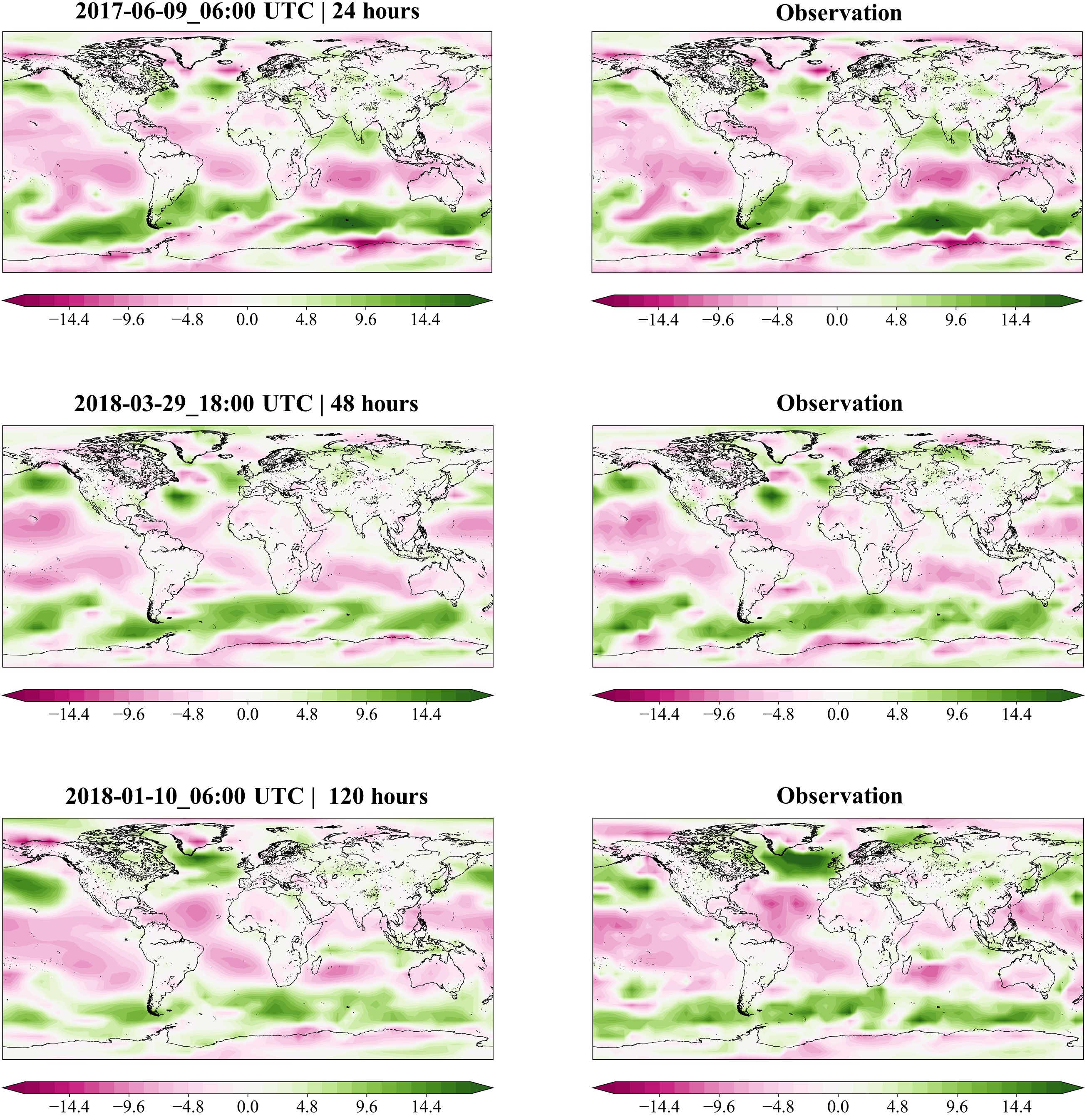}
	\caption{\textbf{Prediction visualization of u10.}}
\label{visual_u10}
\end{figure*}

\begin{figure*}[tb]
    \centering
    \includegraphics[width=38pc]{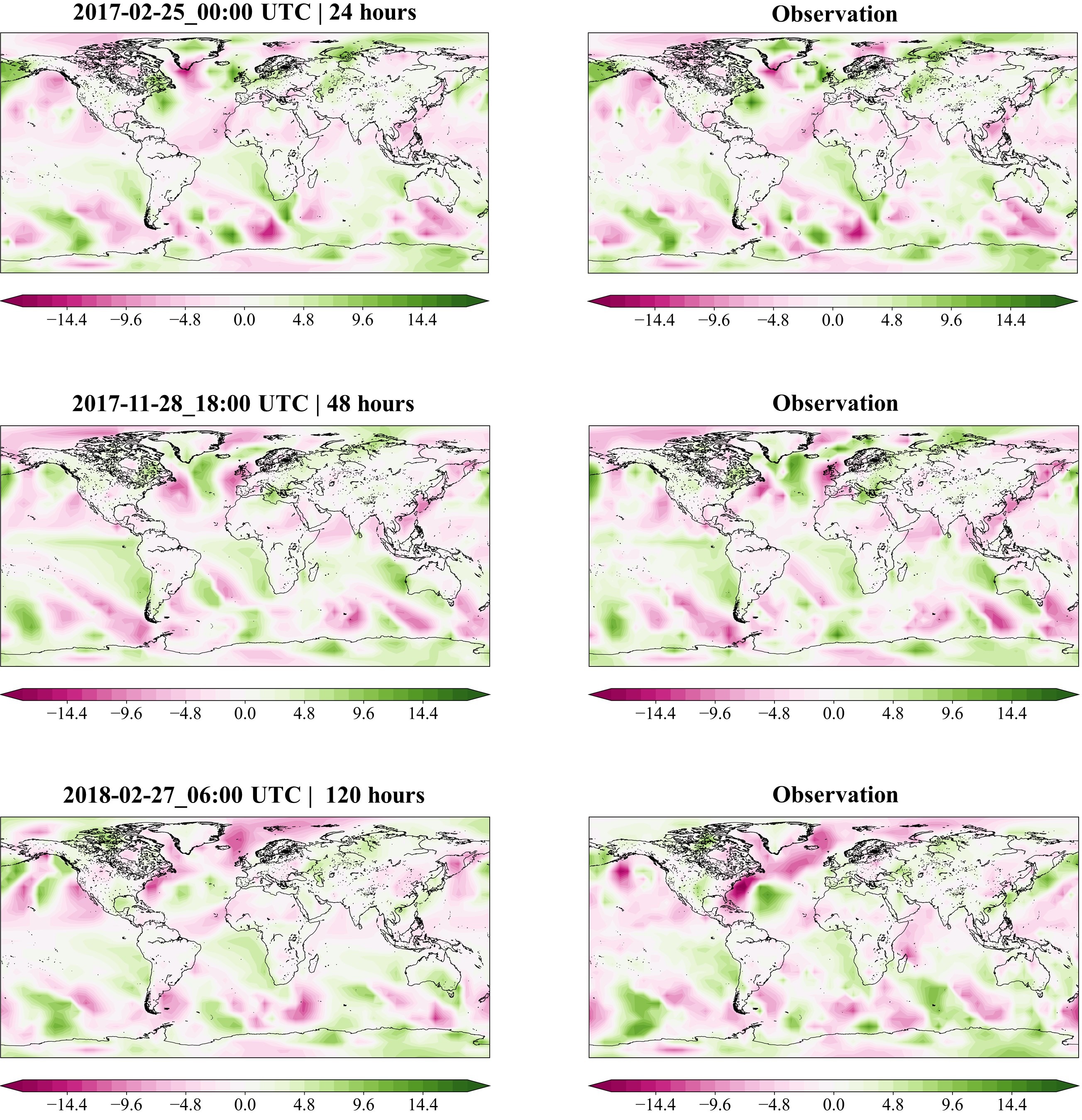}
	\caption{\textbf{Prediction visualization of v10.}}
\label{visual_v10}
\end{figure*}

We present several visualization examples of the predictions generated by PASSAT for t2m (Figure \ref{visual_t2m}), t850 (Figure \ref{visual_t850}), z500 (Figure \ref{visual_z500}), u10 (Figure \ref{visual_u10}), and v10 (Figure \ref{visual_v10}).

\end{document}